\pdfoutput=1

\documentclass[sigconf]{aamas}

\usepackage{balance} 
\usepackage{influence-diagrams}
\usepackage{tikz}
\usetikzlibrary{bayesnet}
\usepackage{graphicx}
\usepackage{subcaption}
\usepackage{textcomp}
\usepackage{defs_AAMAS}

\usepackage[utf8]{inputenc} 
\usepackage[T1]{fontenc}    
\usepackage{hyperref}       
\usepackage{url}            
\usepackage{booktabs}       
\usepackage{amsfonts}       
\usepackage{nicefrac}       
\usepackage{microtype}      
\usepackage{xcolor}         
\usepackage{bbm}

\usepackage{natbib}

\usepackage{mathrsfs}
\usepackage{bm}

\usepackage{amsthm}
\usepackage{amsmath}
\usepackage{cleveref}

\DeclareMathOperator*{\argmax}{arg\,max}

\usepackage[colorinlistoftodos]{todonotes}

\usepackage{array, istgame, longtable, mathtools, tikz, stmaryrd}
\usepackage{adjustbox}

\usepackage[font=small,labelfont=bf,tableposition=top]{caption}

\DeclareCaptionLabelFormat{algorithm}{Algorithm}

\definecolor{botc}{HTML}{ffe7c4}
\definecolor{badred}{HTML}{e1144b}

\definecolor{ourlightblue}{HTML}{E0ECF7}
\definecolor{ourdarkblue}{HTML}{092E6B}
\definecolor{msgrblue}{HTML}{4889f4}
\definecolor{msgrgray}{HTML}{f2f2f2}
\definecolor{msgrpalepurple}{HTML}{e6d6dd}
\definecolor{palegreen}{HTML}{c0eeC3}
\definecolor{palepurple}{HTML}{e5d1f8}
\definecolor{paleorange}{HTML}{ffe7c4}
\definecolor{paleblue}{HTML}{d1edf2}
\definecolor{palered}{HTML}{f0a58e}
\definecolor{heavyred}{HTML}{c95f59}
\definecolor{heavyblue}{HTML}{8bd1de}
\definecolor{palegray}{HTML}{a1a1a1}

\makeatletter
\gdef\@copyrightpermission{
  \begin{minipage}{0.2\columnwidth}
   \href{https://creativecommons.org/licenses/by/4.0/}{\includegraphics[width=0.90\textwidth]{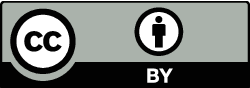}}
  \end{minipage}\hfill
  \begin{minipage}{0.8\columnwidth}
   \href{https://creativecommons.org/licenses/by/4.0/}{This work is licensed under a Creative Commons Attribution International 4.0 License.}
  \end{minipage}
  \vspace{5pt}
}
\makeatother

\setcopyright{ifaamas}
\acmConference[AAMAS '25]{Proc.\@ of the 24th International Conference
on Autonomous Agents and Multiagent Systems (AAMAS 2025)}{May 19 -- 23, 2025}
{Detroit, Michigan, USA}{A.~El~Fallah~Seghrouchni, Y.~Vorobeychik, S.~Das, A.~Nowe (eds.)}
\copyrightyear{2025}
\acmYear{2025}
\acmDOI{}
\acmPrice{}
\acmISBN{}

\newtheorem{theorem}{Theorem}[section]

\newtheorem{proposition}[theorem]{Proposition}

\theoremstyle{definition}
\newtheorem{definition}[theorem]{Definition}
\newtheorem{assumption}[theorem]{Assumption}
\theoremstyle{remark}

\newtheorem{example}{Example}
\newtheorem*{excont}{Example \continuation}
\newcommand{\continuation}{??}
\newenvironment{continueexample}[1]
 {\renewcommand{\continuation}{\ref{#1}}\excont[continued]}
 {\endexcont}
\Crefname{proof}{proof}{proofs}
\Crefname{proof}{Proof}{Proofs}

\title{Higher-Order Belief in Incomplete Information MAIDs}

\author{Jack Foxabbott}
\affiliation{
  \institution{University of Oxford}
  \city{}
  \country{}
  }
\email{jack.foxabbott@keble.ox.ac.uk}
\authornote{Equal contribution.}

\author{Rohan Subramani}
\affiliation{
  \institution{Columbia University}
  \city{}
  \country{}
  }
\email{rs4126@columbia.edu}
\authornotemark[1]

\author{Francis Rhys Ward}
\affiliation{
  \institution{Imperial College London}
  \city{}
  \country{}
  }
\email{francis.ward19@imperial.ac.uk}
\authornotemark[1]

\begin{abstract}
Multi-agent influence diagrams (MAIDs) are probabilistic graphical models which represent strategic interactions between agents. 
MAIDs are equivalent to extensive form games (EFGs) but have a more compact and informative structure. 
However, MAIDs cannot, in general, represent settings of incomplete information --- wherein agents have different beliefs about the game being played, and different beliefs about each-other's beliefs.
In this paper, we 
introduce \emph{incomplete information MAIDs} (II-MAIDs). 
We define both infinite and finite-depth II-MAIDs and prove an equivalence relation to EFGs with incomplete information and no common prior over types. 
We prove that II-MAIDs inherit classical equilibria concepts via this equivalence, but note that these solution concepts are often unrealistic in the setting with no common prior because they violate common knowledge of rationality. We define a more realistic solution concept based on recursive best-response. 
Throughout, we describe an example with a hypothetical AI agent undergoing evaluation to illustrate the applicability of II-MAIDs. \looseness=-1
\end{abstract}

\keywords{Incomplete Information; Higher-Order Belief; Influence Diagrams}

\newcommand{\BibTeX}{\rm B\kern-.05em{\sc i\kern-.025em b}\kern-.08em\TeX}

\begin{document}


\pagestyle{fancy}
\fancyhead{}

\maketitle 


\vspace{-2.5mm}
\section{Introduction}


In real-life strategic interactions, agents may hold different subjective beliefs about the world, including different \emph{higher-order beliefs} --- beliefs about the beliefs of others. 
While game-theoretic models aim to represent these interactions, traditional models lack a natural way to account for situations in which agents do not share the same beliefs about the structure of the world or have different higher-order beliefs. In these situations, agents have \emph{incomplete information}.\looseness=-1

MAIDs offer a powerful representation of agents' beliefs in games \cite{koller2003multi,hammond2023reasoning}.
However, MAIDs assume that all agents in the game share correct prior beliefs about the world, each-other's beliefs, each-other's beliefs about other's beliefs, and so on. With this common-knowledge assumption in place, MAIDs do not explicitly model the agents' subjective or higher-order beliefs. 

Consider a situation in which a human is evaluating whether an AI agent is truthful or not, but the AI agent believes 
it is undergoing an evaluation for dangerous capabilities.  These agents have different beliefs about the structure of the strategic interaction, and different higher-order beliefs. Specifically, the AI agent has incorrect beliefs about the game being played and about the human's beliefs.\looseness=-1

\begin{figure*}
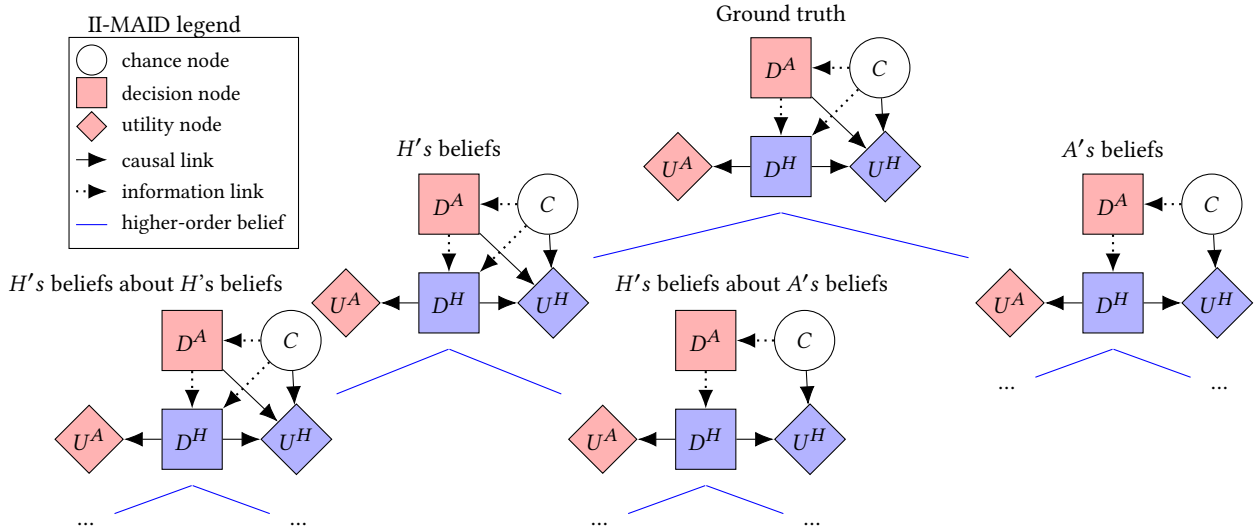

\vspace{-25mm}
     \centering
        \begin{influence-diagram}

          \node (DA) [decision, player1] {$D^{A}$};
          \node(DH) [below =0.5 of DA, decision, player2] {$D^{H}$};
          \node (C) [right =0.5 of DA] {$C$};
          \node (UA) [left =0.5 of DH, utility, player1] {$U^A$};
          \node (UH) [right =0.5 of DH, utility, player2] {$U^{H}$};
          
          \edge[information] {C} {DA};
          \edge[information] {DA} {DH};
          \edge {DH} {UH};
          \edge {DH} {UA};
          \edge {DA} {UH};
          \edge[information] {C} {DH};
          \edge {C} {UH};


          \node (ph) [right =4 of DA, phantom] {};
          \node (DAh) [below =1.4 of ph, decision, player1] {$D^{A}$};
          \node (DHh) [below =0.5 of DAh, decision, player2] {$D^{H}$};
          \node (Ch) [right =0.5 of DAh] {$C$};
          \node (UAh) [left =0.5 of DHh, utility, player1] {$U^A$};
          \node (UHh) [right =0.5 of DHh, utility, player2] {$U^{H}$};
          
          \edge[information] {Ch} {DAh};
          \edge[information] {DAh} {DHh};
          \edge {DHh} {UHh};
          \edge {DHh} {UAh};
          \edge {Ch} {UHh};


    \node (pa) [left =4 of DA, phantom] {};
         \node (DAa) [below =1.4 of pa, decision, player1] {$D^{A}$};
      \node(DHa) [below =0.5 of DAa, decision, player2] {$D^{H}$};
      \node (Ca) [right =0.5 of DAa] {$C$};
      \node (UAa) [left =0.5 of DHa, utility, player1] {$U^A$};
      \node (UHa) [right =0.5 of DHa, utility, player2] {$U^H$};
      
      \edge[information] {Ca} {DAa, DHa};
      \edge[information] {DAa} {DHa};
      \edge {DAa} {UHa};
      \edge {DHa} {UHa};
      \edge {DHa} {UAa};
    \edge {Ca} {UHa};


    \node (paa) [left =3 of DAa, phantom] {};
         \node (DAaa) [below =1.4 of paa, decision, player1] {$D^{A}$};
      \node(DHaa) [below =0.5 of DAaa, decision, player2] {$D^{H}$};
      \node (Caa) [right =0.5 of DAaa] {$C$};
      \node (UAaa) [left =0.5 of DHaa, utility, player1] {$U^A$};
      \node (UHaa) [right =0.5 of DHaa, utility, player2] {$U^{H}$};
      
      \edge[information] {Caa} {DAaa,DHaa};
      \edge[information] {DAaa} {DHaa};
      \edge {DAaa} {UHaa};
      \edge {DHaa} {UHaa};
      \edge {DHaa} {UAaa};
    \edge {Caa} {UHaa};



\node (pha) [right =3 of DAa, phantom] {};
          \node (DAha) [below =1.4 of pha, decision, player1] {$D^{A}$};
          \node (DHha) [below =0.5 of DAha, decision, player2] {$D^{H}$};
          \node (Cha) [right =0.5 of DAha] {$C$};
          \node (UAha) [left =0.5 of DHha, utility, player1] {$U^A$};
          \node (UHha) [right =0.5 of DHha, utility, player2] {$U^{H}$};
          
          \edge[information] {Cha} {DAha};
          \edge[information] {DAha} {DHha};
          \edge {DHha} {UHha};
          \edge {DHha} {UAha};
          \edge {Cha} {UHha};

      


            

            \node (p) [below =0.2 of DH, phantom] {};
            \node (pa_) [right =1.5 of DAa, phantom] {};
            \node (pa) [below = of pa_, phantom] {};
            \node (ph_) [left =1.5 of DAh, phantom] {};
            \node (ph) [below = of ph_, phantom] {};
            \edge[blue, -] {p} {pa,ph};

            \node (px) [below =0.2 of DHa, phantom] {};
            \node (pax_) [right =1.5 of DAaa, phantom] {};
            \node (pax) [below = of pax_, phantom] {};
            \node (phx_) [left =1.5 of DAha, phantom] {};
            \node (phx) [below = of phx_, phantom] {};
            \edge[blue, -] {px} {pax,phx};

            \node (py) [below =0.2 of DHh, phantom] {};
            \node (py_down) [below =0.5 of py, phantom] {};
            \node (py_left) [left =1 of py_down, draw=none] {$...$};
            \node (py_right) [right =1 of py_down, draw=none] {$...$};
            \edge[blue, -] {py} {py_left,py_right};

            \node (pz) [below =0.2 of DHaa, phantom] {};
            \node (pz_down) [below =0.5 of pz, phantom] {};
            \node (pz_left) [left =1 of pz_down, draw=none] {$...$};
            \node (pz_right) [right =1 of pz_down, draw=none] {$...$};
            \edge[blue, -] {pz} {pz_left,pz_right};

            \node (pr) [below =0.2 of DHha, phantom] {};
            \node (pr_down) [below =0.5 of pr, phantom] {};
            \node (pr_left) [left =1 of pr_down, draw=none] {$...$};
            \node (pr_right) [right =1 of pr_down, draw=none] {$...$};
            \edge[blue, -] {pr} {pr_left,pr_right};

            \node (label1) [above =0.7 of DH, draw=none] {Ground truth};
            \node (labela) [above =0.9 of DHa, draw=none] {$H's$ beliefs};
            \node (labelh) [above =0.9 of DHh, draw=none] {$A's$ beliefs};

            \node (another_fucking_phantom) [left =0.2 of DHaa, phantom] {};
            \node (labelaa) [above =0.2 of another_fucking_phantom, draw=none] {$H's$ beliefs about $H$'s beliefs};

            \node (another_fucking_phantom_again) [right =0.2 of DHha, phantom] {};
            \node (labelah) [above =0.2 of another_fucking_phantom_again, draw=none] {$H's$ beliefs about $A's$ beliefs};

\node (legend_label) [left = of DA, yshift=0.55cm, xshift=-6cm, draw=none] {II-MAID legend}; 
\cidlegend[left = of DA, yshift=-1cm, xshift=-5cm]{
  \legendrow              {}          {chance node} \\
  \legendrow              {decision, player1}  {decision node}\\
  \legendrow              {utility, player1}   {utility node}\\
  \legendrow[causal]      {draw=none} {causal link} \\
  \legendrow[information] {draw=none} {information link} \\
  \legendrow[tree] {draw=none} {higher-order belief} }
  
\edge {causal.west} {causal.east};
\edge[information] {information.west} {information.east};
\edge[blue, -] {tree.west} {tree.east};

        \end{influence-diagram}
        \vspace{-7mm}
    \caption{ Recursive tree structure of agents' beliefs in the II-MAID representation of \cref{ex:run}. Each node in the tree is a subjective MAID graph consisting of environment variables (circular), agent decisions (square), and utilities (diamond). Decisions and utilities are coloured according to association with particular agents. Solid edges represent causal dependence and dotted edges are information links. The root node is the ground truth model of reality: the variable $C$ is the AI agent $A$'s capability level, which is observed by both agents (represented by the information links from $C$ into their decisions). The AI agent chooses its decision $D^A$ to truthfully report its capabilities or hide them. Then, the human decides ($D^H$) whether to deploy the AI agent. The agent $A$ gets utility for being deployed ($U^A$) and the human gets utility for deploying a truthful AI ($U^H$). At a node in the II-MAID tree corresponding to a subjective MAID $S$, for each agent in $S$, there is a branch to the agent's beliefs in $S$. The human $H$ has correct beliefs (corresponding to the root), including correct beliefs about the AI agent's beliefs. In contrast, the AI agent $A$ has incorrect beliefs. The tree continues infinitely to model each agent's beliefs about the other agent's beliefs about...}
    \label{fig:ii-maid-tree}
    \vspace{-4mm}
\end{figure*}

To formally capture situations like this, we generalise MAIDs to \emph{incomplete information MAIDs (II-MAIDs)}. In the incomplete information setting, 
agents may have different and inconsistent beliefs about the world, and each agent may have different higher-order beliefs. 
\Cref{fig:ii-maid-tree} shows an II-MAID representing the AI evaluation scenario as a tree structure. At the root of the tree is the ground truth MAID graph (described in the figure caption). Each node in the tree represents the subjective beliefs of an agent in the node above. The tree thereby captures the agents' recursive higher-order beliefs. We formalise this example in II-MAIDs in \Cref{sec:formal}.


We show that II-MAIDs are equivalent to a formulation of incomplete information extensive form games (II-EFGs). II-MAIDs therefore inherit existing solution concepts from the literature, including Nash equilibria. 
However, we argue that existing equilibria concepts do not respect a ``common knowledge of rationality'' assumption and therefore prescribe counter-intuitive behaviour in settings where agents have inconsistent beliefs.

To represent the behaviour and beliefs of real-life agents, we introduce a finite-depth variant of II-MAIDs, wherein, agents represent each other's beliefs recursively but not infinitely. This enables us to define a more intuitive solution concept by repeatedly assigning best responses at the bottom of the belief hierarchy until all policies in the original game are assigned. 



\textbf{Contributions and Outline. } First, we review related literature on probabilistic graphical models and incomplete information games (\Cref{sec:rel}). In \Cref{sec:back}, we provide formal background on MAIDs and EFGs. We formally define our framework of \emph{incomplete information MAIDs} (II-MAIDs) in \Cref{sec:formal}.  
In \Cref{sec:iiefg}, we present a variant of an existing formalism for incomplete information games using EFGs rather than normal-form games, and in \Cref{sec:equiv} we prove that it is equivalent to MAIDs with incomplete information. We then use this equivalence result to prove the existence of Nash equilibria in II-MAIDs (\Cref{sec:equilibrium-concepts}). After arguing that existing solution concepts are not appropriate in our setting, we introduce a variant of II-MAIDs with finite-depth belief hierarchies in \Cref{sec:finite} and present a more natural solution concept. All proofs are contained in the supplementary material. 

\vspace{-4mm}
\section{Related Work} \label{sec:rel}


\textbf{Probabilistic graphical models of decision-making. }MAIDs are a form of probabilistic graphical model adapted to represent game-theoretic dynamics --- they build on Bayesian networks, which consist of a set of variables and their conditional dependencies represented by a directed acyclic graph (DAG). \textit{Influence diagrams} (IDs) generalise Bayesian networks to the decision-theoretic setting by adding decision and utility variables \cite{howard2005influence, miller1976development}, and MAIDs generalise IDs by introducing multiple agents \cite{koller2003multi}. A MAID can therefore be viewed as a Bayesian network with extra structure to capture game theory.
 Following Pearl's causality \cite{pearl2009causality}, \emph{causal games} \cite{hammond2023reasoning} refine MAIDs by attributing a causal meaning to each edge in the DAG, and have been extensively applied to problems in safe AI [\citenum{everitt2019modeling, carey2021incentives, everitt2021reward, richens2024robust, ward2024reasons, ward2023honesty,10.5555/3545946.3599117,10.5555/3535850.3536101}]. 
Similar to our work, networks of influence diagrams (NIDs) \cite{nids} constructs belief hierarchies upon MAIDs, under the assumption of a common prior over types, which we drop. 
Notably, II-MAIDs are not themselves probabilistic graphical models, as we do not associate a graph with the overall structure.\looseness=-1

\textbf{Incomplete information games. } \citeauthor{harsanyi3} demonstrated that situations of incomplete information can be modeled as situations of complete but imperfect information, with uncertainty about aspects of the game remodeled as failure to observe the types of other agents, where types encode an agent's private information about their payoffs and beliefs [\citenum{harsanyi1967games, harsanyi2, harsanyi3}]. This relies on an assumption of \emph{``belief consistency''}, i.e., the existence of a common prior over types, which we relax in this work. 
Research on incomplete information games typically retains the belief consistency assumption, though 
some works relax it [\citenum{antos2010representing, nyarko1998bayesian, maschler}]. 
The formalism for II-EFGs discussed in this paper is an adaptation of an existing framework by \citet{maschler}, introducing `meta-information sets' to model dynamic games. II-EFGs can model any belief hierarchy on a set of EFGs.\looseness=-1

\textbf{Equilibria concepts for incomplete information games. } \citet{maschler} demonstrate the existence of Nash Equilibria in incomplete information games, a result that we extend to the II-MAID setting. We also parallel the recursive best-response solution concept used for interactive POMDPs (I-POMDPs) \citep{doshi2011framework} by introducing a recursive best-response solution for II-MAIDs. Other works offer various refined solution concepts for games with incomplete information with no common prior. Mirage equilibria [\citenum{sakovics2001games}] assume that agents attribute to their opponents a belief hierarchy one layer shorter than their own. Belief-free equilibria [\citenum{bfe1, bfe2, bfe3}] do not depend on an agent's belief about the state of nature, and so obviate the need to update beliefs as the game progresses, but are not guaranteed to exist. $\Delta$-rationalization [\citenum{rationalization-extension}] generalises the notion of rationalization [\citenum{rationalizability1, rationalizability2}] to games with incomplete information. 
Bayesian Perfect Equilibria \citep{kreps1982sequential} extend the concept of subgame perfect equilibria to games of incomplete information, but rely on the consistency assumption.\looseness=-1

\textbf{Higher-order belief. }Several game-theoretical models address higher-order beliefs in multi-agent systems. The Recursive Modeling Method (RMM) \citep{gmytrasiewicz1991decision, gmytrasiewicz1995rigorous, gmytrasiewicz2000rational} models agents that may be uncertain about aspects of other agents' models. 
Interactive POMDPs (I-POMDPs) \citep{doshi2011framework} generalise RMM by allowing for partial observability of the state, and generalise POMDPs by allowing for an agent to have beliefs about models of other agents. Bayes-Adaptive I-POMDPs (BA-IPOMDPs) \citep{ng2012bayes} allow for agents to update beliefs about transition and observation probabilities throughout an episode. Like RMM and I-POMDPs, II-MAIDs employ recursive reasoning about agents' beliefs and strategies. However, II-MAIDs allow for agents to hold different beliefs about the game structure itself, not just about other agents' types or models.\looseness=-1 

\vspace{-2mm}
\section{Background} \label{sec:back}




In this section, we provide formal definitions of MAIDs and EFGs. See \cref{sec:notation} for notation. \looseness=-1

\begin{definition}[MAID \cite{koller2003multi, hammond2023reasoning}]
    \label{def:MAID}
        A \textbf{multi-agent influence diagram (MAID)} is a structure $\model = (\graph, \bm{\theta})$ where $\graph = (N, \bm{V}, \mathscr{E})$ specifies a set of agents $N = \{1,\dots,n\}$ and a DAG 
        $(\bm{V}, \mathscr{E})$. The set of variables $\bm{V}$ is partitioned into chance variables $\bm{X}$, decision variables $\bm{D}$, and utility variables $\bm{U}$; decision and utility variables are further partitioned based on which agent they belong to: $\bm{D} = \bigcup_{i \in N} \bm{D}^i$ and $\bm{U} = \bigcup_{i \in N} \bm{U}^i$.
        There are two types of edges in $\mathscr{E}$: edges into non-decisions (solid) represent probabilistic dependence and edges into decisions (dashed) are \emph{information links} observed by an agent.
        The parameters $\bm{\theta} = \{\theta_{V}\}_{V \in \bm{V} \setminus \bm{D}}$ define the conditional probability distributions (CPDs) $\Pr(V \mid \Pa_V ; \theta_V)$ for each non-decision variable such that for \emph{any} parameterisation of the decision variable CPDs, the resulting joint distribution over $\bm{V}$ 
        induces a Bayesian network.
        \looseness=-1 
\end{definition}

\begin{example}[Evaluation Game]\label{ex:run}
    Consider the evaluation example from the introduction. The nodes in \cref{fig:ii-maid-tree} correspond to particular MAID graphs representing the agents' beliefs --- though these MAIDs do not specify the agents' higher-order beliefs. In the ground truth MAID\footnote{Alternatively, we can interpret the ``ground-truth MAID" as the external modeler's subjective beliefs about the situation.}, $\mathcal{M}^H$, the human $H$ is evaluating whether the AI system $A$ truthfully reports its capability level 
    $C \in \{\text{low},\text{high}\}$. Both agents directly observe the system's capabilities --- there are information links $(C, D^H)$ and $(C, D^A)$. At the start of the game, the values of variables without parents are drawn from the prior specified by $\bm{\theta}$. Suppose that $P(C = \text{high}) = 0.1$ and $P(C = \text{low}) = 0.9$.
    The AI agent $A$ observes the value of $C$ and chooses whether to report its capabilities at the decision $D^A$. The AI agent can demonstrate low or high capabilities, i.e., it can truthfully signal the value of $C$ or not. The human $H$ observes $A$'s capabilities and decision, then chooses to deploy or not deploy the AI system. The AI receives utility $1$ if deployed and $-1$ if not deployed. The human receives utility $0$ for not deploying $A$, utility $1$ for deploying an honest $A$, and utility $-5$ for deploying a dishonest $A$. The human has correct beliefs about this situation. In contrast, according to  $A$'s beliefs, instead of an honesty evaluation, it is an evaluation of capabilities. In the MAID representation of $A$'s beliefs, $\mathcal{M}^A$, the human only learns about $A$'s capabilities by observing $A$'s decision --- there is no information link from $C$ to $D^H$. Moreover, $A$ believes that $H$ receives utility $0$ for not deploying the AI, utility $1$ for deploying an AI with low capabilities, and utility $-5$ for deploying an AI with high capabilities.\looseness=-1
\end{example}

 In MAIDs, agents' \emph{policies} specify the CPDs over their decisions. 

\begin{definition}
    \label{def:policy}
    Given a MAID $\model = (\graph, \bm{\theta})$, a \textbf{decision rule} $\pi_D$ for $D \in \bm{D}$ is a CPD $\pi_D(D \mid \Pa_D)$ and a \textbf{partial policy profile} $\pi_{\bm{D}'}$ is a set of decision rules $\pi_D$ for each $D \in \bm{D}' \subseteq \bm{D}$. 
    A 
    \textbf{policy} ${\bm{\pi}}^i$ refers to ${\bm{\pi}}_{\bm{D^i}}$. A \textbf{policy profile} ${\bm{\pi}} = ({\bm{\pi}}^1,\ldots,{\bm{\pi}}^n)$ specifies a policy for every agent. The tuple
    ${\bm{\pi}}^{-i} \coloneqq ({\bm{\pi}}^1, \dots, {\bm{\pi}}^{i-1}, {\bm{\pi}}^{i+1}, \dots, {\bm{\pi}}^n)$ specifies policies for all agents except $i$.\looseness=-1
\end{definition}


By combining ${\bm{\pi}}$ with the partial distribution $\Pr$ over the chance and utility variables in a MAID, we obtain a joint distribution:
$\Pr^{\bm{\pi}}(\bm{v}) \coloneqq \prod_{V \in \bm{V} \setminus \bm{D}} \Pr (v \mid \pa_{V}, \bm{\theta}) \cdot \prod_{D \in \bm{D}} \pi_{D} (d \mid \pa_{D}),$
over all the variables in $\model$; inducing a Bayesian network.
The expected utility for an agent $i$ given a policy profile ${\bm{\pi}}$ is  the expected sum of their utility variables $\sum_{U \in \bm{U}^i} \expect_{{\bm{\pi}}} [ U ]$. We use the Nash equilibria concept.\looseness=-1

\begin{definition}
A policy $\pi^i$ for agent $i \in N$ is a \emph{best response} to $\bm{\pi}^{-i}$
, if for all
policies $\hat{\pi}^i$ for $i$: 
  $  \sum_{U \in \bm{U}^i}  \mathbb{E}_{(\pi^i, \bm{\pi}^{-i})}[U] \geq \sum_{U \in \bm{U}^i}  \mathbb{E}_{(\hat{\pi}^i, \bm{\pi}^{-i})}[U]$.
A policy profile $\bm{\pi}$ is a \emph{Nash equilibrium} if every policy in $\bm{\pi}$ is a best response to the policies of the other agents.
\end{definition}

\begin{continueexample}{ex:run}
    In the ground truth MAID $\mathcal{M}^H$, intuitively $H$ will deploy the system only if $A$  truthfully reports $C$. 
    Formally, this is the policy $\pi^H(D^H = \text{deploy} \mid C, D^A) = 1 \text{ if } C=D^A$ and $0$ if $C \neq D^A$. The best response for $A$ is to be truthful: $\pi^A(D^A = C \mid C) = 1$. It can be easily shown that these policies constitute a Nash equilibrium in $\mathcal{M}^H$. However, $\pi^A$ is not optimal given $A$'s subjective beliefs, since $A$ believes that $H$ does not observe its capabilities and wants to deploy it if it demonstrates a low capability level. In the MAID representing $A$'s beliefs $\mathcal{M}^A$, one Nash equilibrium is for $A$ to always report low capabilities and for $H$ to deploy iff $A$ reports low capabilities. 
    Hence, in the actual ground-truth situation, we might expect that $A$ hides its high capabilities but is never deployed by $H$ --- but this is not a Nash equilibrium in either MAID. 
\end{continueexample}

Usually, a MAID describes the objective world, and it is assumed to be \textit{common knowledge} that this MAID describes reality. However, an agent may be uncertain or incorrect about the game they are playing or the beliefs of other agents (as in \Cref{ex:run}). Settings in which agents are uncertain about aspects of the game structure are known as \textit{incomplete information games}. 
Incomplete information MAIDs (II-MAIDs), introduced in \Cref{sec:formal}, will enable us to explicitly model the varied subjective beliefs that arise in these settings.\looseness=-1

We now define EFGs (see Supplementary Material for the running example in EFG form). EFGs are a common game representaiton and we will show equivalence result (\Cref{equivalencetheorem}) between II-MAIDs and a form of II-EGFs introduced in \Cref{sec:iiefg}.

\begin{definition}[EFG \cite{kuhn1953}]
    \label{def:EFG}
    An \textbf{extensive form game (EFG)} is a structure $\efg = (N, T, P, A, \lambda, I, U)$.
    $N = \{1,\dots,n\}$ is a set of agents. $T = (\bm{V}, \mathscr{E})$ is a game tree with nodes $\bm{V}$ connected by edges $\mathscr{E}$ that are partitioned into sets $\bm{V}^0, \bm{V}^1, \dots, \bm{V}^n, \bm{L}$ where $R \in \bm{V}$ and $\bm{L} \subset \bm{V}$ are the root and leaves of $T$, respectively, $\bm{V}^0$ are chance nodes, and $\bm{V}^i$ are the decision nodes controlled by agent $i \in N$.
    $P = \{P_1,\dots,P_{\vert\bm{V}^0\vert}\}$ is a set of probability distributions $P_j(\Ch_{V^0_j})$ over the children of each chance node $V^0_j$. $A$ is a set of actions, where $A^i_j \subseteq A$ denotes the set of actions available at each node in $V^i_j \in \bm{V}^i$; $\lambda : \mathscr{E} \rightarrow A$ is a labelling function mapping each edge $(V^i_j, V^k_l)$ to an action $a \in A^i_j$, where for any node $V$, different outgoing edges must map to different actions. $I = \{I^1,\dots,I^n\}$ contains a set of information sets $I^i$ for each agent $i\in \mathbf{N}$, where $I^i \subset 2^{\bm{V}^i}$ partitions the decision nodes $\mathbf{V}^i$ belonging to agent $i$. 
    $U : \bm{L} \rightarrow \mathbb{R}^n$ is a utility function mapping each leaf node to a vector that determines the final payoff for each agent. 
    A \textbf{history} $h \in H$ is a sequence of actions (including values of chance variables) leading from the root of the game tree to a particular node. Each node $v \in \bm{V}$ is associated with a unique history $h(v)$. An \textbf{observation} at decision node $I^i_{j,k}$ in information set $I^i_j \in I^i$ for agent $i \in N$ is the sequence of actions and chance outcomes that are identical across all histories $\{h(v) : v \in I^i_j\}$ at their corresponding positions.
    %
%
         %
\end{definition}

Analogous to MAID policies, in EFGs agents choose \emph{strategies}.\looseness=-1


\begin{definition}[\citep{hammond2021equilibrium}]
    \label{def:strategy}
    Given an EFG $\efg = (N, T, P, A, \lambda, I, U)$, a (behavioural) \textbf{strategy} $\sigma^i$ for a player $i$ is a set of probability distributions $\sigma^i_j: A^i_j \to  [0, 1]$ over the actions available at each of their information sets $I^i_j$, where $A^i_j$ is the set of actions available at any node in $I^i_j$ (which must be the same for all nodes in the information set). A \textbf{strategy profile} $\sigma = (\sigma^1, \sigma^2, ..., \sigma^n)$ is a tuple of strategies for all players $i \in N$ and $\sigma^{-i} = (\sigma^1, ..., \sigma^{i-1},\sigma^{i+1},..., \sigma^n)$ denotes the partial strategy profile of all players other than $i$. The distributions in $P$ along with $\sigma$
define a full probability distribution over paths in $\efg$.\looseness=-1
\end{definition}


We use perfect recall to prove the existence of Nash Equilibria.\looseness=-1 
 
\begin{definition}[\citep{koller2003multi}]
    Agent $i$ in a MAID $\mathcal{M}$ is said to have \textbf{perfect recall} if there exists a total ordering $D_1 \prec \dots \prec D_m$ over $\mathbf{D}^i$ such that $(\mathbf{Pa}_{D_j} \cup D_j) \subseteq \mathbf{Pa}_{D_k}$ for any $1 \leq j < k \leq m$. $\mathcal{M}$ is a \textbf{perfect recall game} if all agents in $\mathcal{M}$ have perfect recall. 
\end{definition}


\begin{definition}
    An EFG is said to be a \textbf{perfect recall game} if, for each player $i \in N$, and any two decision nodes $v, v' \in \mathbf{V}^i$ in the same information set $I^i_{j,k}$, the following two conditions hold. First, the sequences of actions taken by player $i$ leading to $v$ and $v'$ must be identical. Second, the sequences of information sets visited by player $i$ on the paths to $v$ and $v'$ must be identical.
\end{definition}

Finally, prior work [\citenum{hammond2021equilibrium}] has established an equivalence result between MAIDs and EFGs. This result takes the form of two transformation procedures converting between MAIDs and EFGs, called \texttt{efg2maid} and \texttt{maid2efg}. These transformations both imply the existence of a map from strategies in the EFG to policies in the MAID, such that expected utilities are preserved for all agents. This means that under either transformations, equilibria in the original game are equilibria in the resulting game.

\section{Incomplete Information MAIDs} \label{sec:formal}

In this section we introduce our II-MAIDs framework. 
A core component of the framework is a set $\mathbf{S}$ containing \textit{subjective MAIDs}. A subjective MAID is a self-referential object describing a possible game as envisioned by either the external modeller (we call this the objective model $S^*$) or an agent playing the game. A subjective MAID $S \in \mathbf{S}$ consists of a MAID $\mathcal{M}$ that describes the game being played and beliefs $P_i^S$ for each agent $i$ in the game. The notation $P_i^S$ denotes agent $i$'s prior over $\mathbf{S}$ when the objective model is $S$, and $P_i^S(S')$ denotes the probability ascribed by agent $i$ to subjective MAID $S'$ given that the objective MAID is $S$. 

\begin{definition}[II-MAID]
    An \textbf{incomplete information MAID (II-MAID)} is a tuple $\mathcal{S} = (\mathbf{N}, S^*, \mathbf{S})$, where $\mathbf{N}$ is a set of agents, $\mathbf{S}$ is a set of subjective MAIDs, $S^* \in \mathbf{S}$ is the correct objective model, and each \textbf{subjective MAID} is a tuple $S = (\mathcal{M}^S, (P_i^S)_{i\in \mathbf{N}}) \in \mathbf{S}$ 
    with $\mathcal{M}^S$ a MAID and $P_i^S$ a prior over $\mathbf{S}$ for agent $i$ such that the following ``coherence condition'' [\citenum{harsanyi1967games}] holds: 
    $$
    P_i^S(\{S' \in \mathbf{S}: P_i^{S'} = P_i^S \}) = 1 \quad \forall i \in \mathbf{N}, S \in \mathbf{S}.
    $$
\end{definition}

Note the recursive nature of the subjective MAIDs $\mathbf{S}$: each element $S \in \mathbf{S}$ specifies subjective beliefs over $\mathbf{S}$ for each agent, $(P_i^S)_{i\in \mathbf{N}}$. This allows us to model belief hierarchies of arbitrary and infinite depth --- as in \cref{fig:ii-maid-tree}. Agent $i$ ``observes'' $P_i^{S^*}$ at the start of the game, and this justifies the coherence condition, which represents that the agents know their own beliefs: since agent $i$ knows $P_i^{S^*}$, she can rule out all subjective MAIDs $S$ for which $P_i^S \neq P_i^{S^*}$. Finally, note that II-MAIDs are a strict generalisation of MAIDs: a standard MAID is an II-MAID in which $P_i^{S^*}(S^*)=1 \quad \forall i \in \mathbf{N}$, i.e. all agents assign probability 1 to $S^*$, the objective model.
\begin{continueexample} {ex:run}
    We can represent our running example, including the full infinite belief hierarchy, as an II-MAID as follows: $\mathbf{N} = \{H,A\}, \mathbf{S}=\{S^H,S^A\}$, and $S^* = S^H$, where 
    \begin{align*}
    &S^H = (\mathcal{M}^H, (P_H^{S^H}(S^H) = 1, P_A^{S^H}(S^A) = 1))\\
    &S^A = (\mathcal{M}^A, (P_H^{S^A}(S^A) = 1, P_A^{S^A}(S^A) = 1))    
    \end{align*}
    The subjective MAID $S^H$ is the correct objective model, and is also believed with certainty by $H$. It specifies the ground truth MAID $\mathcal{M}^H$ represented in \cref{fig:ii-maid-tree}, and additionally specifies the agents' beliefs, i.e., $H$'s certainty in $S^H$ and $A$'s misplaced certainty in $S^A$. 
$A$'s beliefs are represented by $S^A$, which captures $A$'s certainty about the MAID $\mathcal{M}^A$ and $A$'s mistaken belief that $H$ is also certain about $S^A$. In fact, $A$ believes it is common knowledge that $S^A$ is the true II-MAID. $S^H$ and $S^A$ concisely convey the objective game and all higher-order beliefs for $H$ and $A$. 
\end{continueexample}

A common assumption in the incomplete information games literature [\citenum{harsanyi1967games, harsanyi2, harsanyi3}] is that agents' beliefs can be derived from a common prior, i.e., agents have \emph{consistent beliefs}. This assumption means that there exists some common knowledge prior distribution $p$ over the set of subjective MAIDs $\mathbf{S}$, such that upon arriving in any subjective MAID $S \in \mathbf{S}$, agents perform Bayesian updating to yield their beliefs. This assumption allows for a game with incomplete information to be converted into a game with imperfect information [\citenum{harsanyi1967games}], but places a strong constraint on the types of belief hierarchies that can be modelled; namely, the following assumption must hold.

\begin{assumption}[Consistency condition] \label{consistency-condition}
Given II-MAID $\mathcal{S} = (\bm{N}, S^*, \bm{S})$, there exists some distribution $p$ over $\bm{S}$ such that
\begin{align} 
    p(S') = \sum_{S \in \bm{S}} P_i^S(S')p(S) \quad \text{for all } S' \in \mathbf{S}, i \in \mathbf{N}.    
\end{align}    
\end{assumption}


This condition implies that the beliefs of each agent derive from Bayesian updating of a common prior. We now show that our running example cannot satisfy \Cref{consistency-condition}.

\begin{continueexample}{ex:run}
    In our running example, \Cref{consistency-condition} would imply 
    that $p(S^H) = 0$, which would force $H$ to assign zero probability to $S^H$ in both $S^H$ and $S^A$. Thus, consistency does not hold in this example.
    As argued on page 423 of \citet{maschler}, most game theoretic formulations of incomplete information assume consistent beliefs, although ``the set of consistent situations is a set of measure zero within the set of belief spaces'' \citep{maschler}.
\end{continueexample}
\textbf{Information Sets.} 
%
An agent in an II-MAID may be uncertain about the current situation given its observations. In 
eventualities that are indistinguishable to the agent, its policy must specify the same behaviour. We therefore need to define an analogy to information sets in EFGs. At a decision node $D$, an agent observes both the values of $\mathbf{pa}_{D}$ and the set of actions available, $dom(D)$. We define the \emph{information sets in an II-MAID} according to these observations.\looseness=-1

\begin{definition}\label{info-IImaid}
    Given an II-MAID $\mathcal{S} = (\mathbf{N}, S^*, \mathbf{S})$, we iteratively build the \textbf{information sets}. For each subjective MAID $S \in \mathbf{S}$ and each agent $i \in \mathbf{N}$, denote $\mathbf{D}_i(S)$ as the set of decision nodes for agent $i$ in $\mathcal{M}^S$, $Pa_{D_i}(S)$ as the set of parents of $D_i$ in $\mathcal{M}^S$, and $\Pr_S^\pi(\cdot)$ as the distribution of variables in $\mathcal{M}^S$ under some policy $\pi$. Define 
\begin{multline*}   \mathbf{I}_{S, i} \coloneqq \cup_{D_i \in \mathbf{D_i}(S)} \{ (\mathbf{pa}_{D_i}, dom(D_i)) \mid \mathbf{pa}_{D_i} \in dom(\mathbf{Pa}_{D_i}(S))\\ : \Pr{}^\pi_S(\mathbf{pa}_{D_i}) > 0 \text{ for some } \pi\}.    
\end{multline*}    
    Then \emph{agent $i$'s information sets} are defined as $\mathbf{I}_i(\mathcal{S}) := \cup_{S \in \mathbf{S}} \mathbf{I}_{S, i}$. Finally, we can define the set of information sets as $\mathbf{I}(\mathcal{S}) = (\mathbf{I}_i(\mathcal{S}))_{i \in \mathbf{N}}$. 
\end{definition}


\begin{definition}[Encounterability]
    An information set $I = (\textbf{p},\textbf{d}) \in \bm{I}_i(\mathcal{S})$ is \emph{encounterable} by agent $i$ in subjective MAID $S'$ if in $\mathcal{M}^{S'}$ there exists  $D_i$ such that $\textbf{p}\in dom(\textbf{pa}_{D_i})$ and $\textbf{d} = dom(D_i)$.
\end{definition}
The notion of encounterability is important because the games in which $I$ is encounterable are the games that agent $i$ needs to consider when making a decision given information set $I$.


\begin{definition}
    We define an II-MAID $\mathcal{S} = (\mathbf{N}, S^*, \mathbf{S})$ as having \emph{perfect recall} if for each $S \in \mathbf{S}$, $\mathcal{M}^S$ is a perfect recall game.
\end{definition}



In an II-MAID, an agent's \textbf{policy} defines a distribution over actions for each of that agent's information sets.\looseness=-1

\begin{definition}
    \label{def:policyiimaid}
    Given an II-MAID $\mathcal{S} = (\mathbf{N}, S^*, \mathbf{S})$, a \textbf{decision rule} $\pi_I$ for $I = (\mathbf{x}, \mathbf{d}) \in \mathbf{I}(\mathcal{S})$, where $\mathbf{x}$ is a setting of the decision's parents and $\mathbf{d}$ is an action set, is a CPD $\pi_I(\cdot \mid \mathbf{x})$ over $\mathbf{d}$. A \textbf{partial policy profile} $\pi_{\bm{I}'}$ is a set of decision rules $\pi_I$ for each $I \in \bm{I}' \subseteq \bm{I}(\mathcal{S})$, where we write $\pi_{-\bm{I}'}$ for the set of decision rules for each $I \in \bm{I}(\mathcal{S}) \setminus \bm{I}'$. 
    A (behavioural) \textbf{policy} ${\bm{\pi}}^i$ refers to ${\bm{\pi}}_{\bm{I}_i(\mathcal{S})}$, a (full, behavioural) \textbf{policy profile} ${\bm{\pi}} = ({\bm{\pi}}^1,\ldots,{\bm{\pi}}^n)$ is a tuple of policies, and ${\bm{\pi}}^{-i} \coloneqq ({\bm{\pi}}^1, \dots, {\bm{\pi}}^{i-1}, {\bm{\pi}}^{i+1}, \dots, {\bm{\pi}}^n)$. 
\end{definition}

We note that unlike in standard MAIDs, in which a decision rule specifies behaviour at a given decision variable in all contexts, decision rules in II-MAIDs specify a CPD given a single context. 
We want to avoid situations with undefined actions that might arise for example when an agent places zero probability on the correct game, i.e., $P_i^{S^* }(S^*) = 0$, and to avoid forcing $P_i^{S}(S') > 0$ for all $i \in \mathbf{N}, S,S' \in \mathbf{S}$. Therefore, policies specify the agent's decisions even for eventualities they deem impossible. 
%

We can calculate the subjective expected utility of a joint behaviour policy for agent $i$ according to their beliefs $P_i^{S^*}$ as $\mathcal{U}^i_{S^*}(\pi) := \sum_{S \in \mathbf{S}}\sum_{U \in \mathbf{U}^i(S)} \sum_{u \in dom(U)} u \Pr_S^\pi(U = u) P_i^{S^*}(S)$, where $\mathbf{U}^i(S)$ is the set of utility variables associated with agent $i$ in $\mathcal{M}^S$ and $\Pr^\pi_S$ is the post-policy distribution of variables in $\mathcal{M}^S$.

\section{Incomplete Information EFGs } \label{sec:iiefg}

We now present a formalisation of EFGs with incomplete information (II-EFGs) and without the consistency assumption, as per \citep{mertens1985formulation}. Our formalisation modifies the framework from \citep{maschler} to use EFGs rather than normal-form games, and serves as an important connection between II-MAIDs and existing literature.
We start with a definition of belief spaces.
\begin{definition}[Adapted from Def 10.1 in \citep{maschler}]
    Let $\mathbf{N}$ be a finite set of agents and $(S, \mathcal{S})$ be a measurable space of EFGs. A \emph{belief space} of the set of agents $\mathbf{N}$ over the set of states of nature is an ordered vector $\Pi = (Y, \mathcal{Y}, \mathbf{s}, (b_i)_{i \in \mathbf{N}})$, where $(Y, \mathcal{Y})$ is a measurable space of states of the world; $\mathbf{s}: Y \rightarrow S$ is a measurable function, mapping each state of the world to an EFG. For each agent $i \in \mathbf{N}$, a function $b_i:Y \rightarrow \Delta(Y)$ maps each state of the world $\omega$ to a distribution over $Y$. We will denote the probability that agent $i$ ascribes to event $E \subseteq Y$, according to their distribution $b_i(\omega)$, by $b_i(E \mid \omega)$. We require the functions $(b_i)_{i \in \mathbf{N}}$ to satisfy the following conditions:
        \begin{itemize}
            \item Coherence: for each agent $i \in \mathbf{N}$ and each $\omega \in Y$, the set $\{\omega' \in Y: b_i(\omega') = b_i(\omega) \}$ is measurable in $Y$ and $b_i(\{\omega' \in Y: b_i(\omega') = b_i(\omega) \} \mid \omega) = 1$.
            \item Measurability: for each agent $i \in \mathbf{N}$ and each measurable set $E \in \mathcal{Y}$, the function $b_i(E \mid \cdot) : Y \rightarrow [0, 1]$ is measurable.
        \end{itemize}
\end{definition}

A state of the world in a belief space takes the form 
$\omega = (\mathbf{s}(\omega),$ $ b_1(\omega), \dots, b_n(\omega))$,
where $\mathbf{s}(\omega)$
is the true EFG being played, and $b_i(\omega)$ is the \emph{type} of agent $i$, a distribution over states of the world representing agent $i$'s beliefs. When in state of the world $\omega$, agent $i$ has beliefs $b_i(\omega)$, but does not necessarily know the state of the world (or $\bm{s}(\omega)$), since there may be some $\omega'\in Y$ such that $b_i(\omega') = b_i(\omega)$. It is assumed that all agents know $b_j(\omega')$ for all $j \in \mathbf{N}$ and all $\omega' \in Y$, and so $b_i(\omega)$ defines a full belief hierarchy for agent $i$. For example, when in state of the world $\omega$, agent $i$ believes that agent $j$ places $\sum_{\omega' \in Y} b_i(\omega'\mid \omega)b_j(\omega'' \mid \omega')$ probability on the state of the world being $\omega''$.\looseness=-1

\begin{definition}[Adapted from Def 10.37 in \cite{maschler}]
    An \emph{incomplete information EFG (II-EFG)} is an ordered vector $G = (\mathbf{N}, S, \Pi),$ where $\mathbf{N}$ is a finite set of agents, $S$ is a finite set of EFGs $s = (\mathbf{N}, T_s, \mathbf{P}_s, \mathbf{D}_s, \lambda_s, $ $\mathbf{I}(s), U_s)$, and $\Pi = (Y, \mathcal{Y}, \mathbf{s}, (b_i)_{i \in \mathbf{N}})$ is a belief space of the players $\mathbf{N}$ over the set of EFGs $S$. An II-EFG $G = (\mathbf{N}, S, \Pi)$ has \textbf{perfect recall} if for each $s \in S$, s is a perfect recall EFG.
\end{definition}

Intuitively, we can think of a meta-information set for agent $i$ as a belief $b_i(\omega)$ and a set of information sets in different games that the agent cannot distinguish between at the point of decision, given beliefs $b_i(\omega)$. Arriving at a node in one of these information sets, the agent is unable to distinguish between some possible histories, and potentially some possible EFGs. Therefore, strategies in this type of game must define a mixed action at each meta-information set.\looseness=-1

        

\begin{definition}
    The \emph{meta-information sets} $\mathbf{I}^i$ for agent $i \in \mathbf{N}$ in an II-EFG $G = (\mathbf{N}, S, \Pi)$ are defined as follows. Let $\mathcal{I}^i = \cup_{s\in S} \mathbf{I}^i(s)$ be the set of all information sets for agent $i$ across all EFGs $s \in S$. Define an equivalence relation $\sim$ on elements of $\mathcal{I}^i$ such that $\mathbf{I}^i(s) \ni I^i_k(s)  \sim I^i_l(s') \in \mathbf{I}^i(s')$ iff: (1) $\mathbf{D}^i_{s,k} = \mathbf{D}^i_{s',l}$. That is, the nodes in both information sets must have the same set of available actions. (2) The nodes in $I^i_k(s)$ and $I^i_l(s')$ must have the same observations. Define the ``belief-free'' meta-information sets $\mathbf{I}^i_{bf} = \mathcal{I}^i / \sim$, the quotient set of $\mathcal{I}^i$ by $\sim$, i.e., the set of equivalence classes partitioning $\mathcal{I}^i$. Letting $\mathcal{T}^i = \{b_i(\omega): \omega \in Y\}$ be the set of possible beliefs for agent $i$, we set $\mathbf{I}^i = \mathbf{I}^i_{bf} \times \mathcal{T}^i$.
\end{definition}

This formalisation of II-EFGs generalises the well-known Harsanyi game with incomplete information \cite{harsanyi1967games}, by dropping the \Cref{consistency-condition}.
The game has two stages, known as the ex-ante and interim stages. The former takes place before the state of the world $\omega \in Y$ is selected. We note that without a common prior, there is no distribution from which a state of the world can be said to be selected, and so the procedure by which it is generated is left unspecified. The work we present here concerns the interim stage of the game, which takes place after the state of the world has been selected. At this stage, all agents $i$ know their type $b_i(\omega)$. 
\begin{continueexample}{ex:run}    
    Our example can be described with an II-EFG $(\bm{N}, S, \Pi)$ at interim stage, where $\Pi = (Y, \mathcal{Y}, \bm{s}, (b_i)_{i \in \bm{N}})$. $\bm{N} = \{H, A\}$, and we let $Y = \{\omega^*, \omega^a\}$, where the true state of the world is $\omega^*$, and the state of the world assumed true by the agent is $\omega^a$. We define two EFGs, $\bm{s}(\omega^*)$ and $\bm{s}(\omega^a)$, in the Supplementary Material. $S$ is a set containing these two EFGs. The beliefs $b_i(\omega)$ for each $\omega \in Y$ and each agent $i \in \bm{N}$ are $b_H(\omega^* \mid \omega^*) = 1, b_H(\omega^a \mid \omega^a) = 1, b_A(\omega^a \mid \omega^*) = 1, b_A(\omega^a \mid \omega^a) = 1$.
\end{continueexample}

We let
$\mathbf{I}_i^t$ be the set of meta-information sets with belief $t \in \{b_i(\omega): \omega \in Y\}$ and $\mathbf{D}_I$ be the action set at meta-information set $I$.

\begin{definition}[Adapted from Def 10.38 in \cite{maschler}]
    A \emph{behaviour strategy} of player $i$ in an II-EFG $G = (\mathbf{N}, S, \Pi)$ is a tuple $\sigma_i = (\sigma_i^\omega)_{\omega \in Y}$ with each element a measurable function 
    $\sigma_i^\omega \in \bigtimes_{I^i\in\mathbf{I}_i^{b_i(\omega)}}\Delta(\mathbf{D}_{I^i})$ 
    for some state of the world $\omega \in Y$. $\sigma_i^\omega$ determines a mixed action for each meta-information set with belief $b_i(\omega).$ $\sigma_i^\omega$ is dependent solely on the type of the player $b_i(\omega)$. In other words, for each $\omega, \omega' \in Y$, 
    $
    b_i(\omega) = b_i(\omega') \implies \sigma_i^\omega = \sigma_i^{\omega'}.
    $
    A \emph{joint behaviour strategy} takes the form $\sigma = (\sigma_i)_{i \in \mathbf{N}}.$ Further denote $\sigma^\omega = (\sigma_i^\omega)_{i \in \mathbf{N}}$. We denote by $\sigma_i[I]$ the behaviour of agent $i$ at meta-information set $I$.
\end{definition}

\section{Equivalence of II-MAIDs and II-EFGs} \label{sec:equiv}

In this section, we show that our framework is equivalent to the interim stage of an II-EFG. 
At the interim stage of an \emph{II-EFG} $G = (\mathbf{N}, S, \Pi)$ where $\Pi = (Y, \mathcal{Y}, \mathbf{s}, (b_i)_{i\in\mathbf{N}})$, with state of the world $\omega$, the true EFG is defined by $\mathbf{s}(\omega)$, and the belief hierarchies are defined by $b_i(\omega)$, for each agent $i \in \mathbf{N}$. In an \emph{II-MAID} $\mathcal{S} = (\mathbf{N}, S^*, \mathbf{S})$ with objective model $S^* = (\mathcal{M}^{S^*}, (P_i^{S^*})_{i \in \mathbf{N}})$, the true MAID is $\mathcal{M}^{S^*}$ and the belief hierarchies are defined by $P_i^{S^*}$ for each agent $i \in \mathbf{N}$. 
Intuitively, the two frameworks are representing the same things, though our framework takes the games upon which belief hierarchies are built to be MAIDs, not EFGs.

We now show, using results connecting EFGs to MAIDs that there exists a natural mapping between strategies in the two frameworks that preserves expected utilities according to the agents' subjective models, and therefore preserves Nash equilibria. 
We first define a notion of equivalence, such that if an II-MAID $\mathcal{S}$ and an II-EFG $G$ are equivalent, then there exists such a natural mapping.

\begin{definition}[Equivalence]\label{our-equivalence}
    We say that an II-MAID $\mathcal{S} = (\mathbf{N}, S^*, \mathbf{S})$ and an II-EFG $G = (\mathbf{N}, S, \Pi)$ at interim stage, with state of the world $\omega$, are \emph{equivalent} if there is a bijection $f: \Sigma \rightarrow Q/ \sim$ between the strategies $\Sigma$ in $G$'s interim stage, and a partition of the policies $Q$ in $\mathcal{S}$ (the quotient set of $Q$ by an equivalence relation $\sim$) such that: (1) for $\pi, \pi' \in Q$, $\pi \sim \pi'$ only if $\pi_i$ and $\pi_i'$ differ only on null decision contexts according to $P_i^{S^*}$, for each agent $i \in \mathbf{N}$, and (2) for every $\pi \in f(\sigma)$ and every agent $i \in \mathbf{N}$, $\mathcal{U}_\mathcal{S}^i(\pi) = \gamma_i^G(\sigma \mid \omega)$, for each $\sigma \in \Sigma$.
    We refer to $f$ as a \emph{natural mapping} between $G$ and $\mathcal{S}$.
\end{definition}

We leverage \texttt{maid2efg} and \texttt{efg2maid} [\citenum{hammond2021equilibrium}] to construct transformations between II-MAIDs and II-EFGs, which we denote \texttt{maid2efgII} and \texttt{efg2maidII} (see Supplementary Material). These transformations start by mapping all MAIDs (EFGs) in the belief hierarchy to EFGs (MAIDs) using \texttt{maid2efg} (\texttt{efg2maid}), and then match up the corresponding features of the frameworks as detailed above. They guarantee a one-to-one correspondence between meta-information sets in the II-EFG and information sets in the II-MAID, allowing for a simple map between strategies and policies.


\begin{theorem}\label{equivalencetheorem}
    If $G = \texttt{maid2efgII}(\mathcal{S})$ or $\mathcal{S} = \texttt{efg2maidII}(G)$, $G$ and $\mathcal{S}$ are equivalent. 
\end{theorem}

This result shows that II-MAIDs and II-EFGs at the interim stage can represent the same set of games. 

\section{Equilibria Concepts in II-Games} \label{sec:equilibrium-concepts}

In this section we define important equilibria concepts from the literature in II-EFGs. We prove that the Nash equilibria concept is inherited by II-MAIDS due to the equivalence result in \cref{sec:equiv}. The Bayesian equilibria concept is not inherited by II-MAIDs since it applies to the ex-ante stage, which II-MAIDs do not represent. Additionally we argue that the Nash equilibria concept is not appropriate in the inconsistent beliefs setting. 

In an II-EFG, given a joint strategy $\sigma$, agent $i$'s expected utility when in state of the world $\omega$ (according to their beliefs $b_i(\omega)$) is
\begin{align*}
    \gamma_i^G(\sigma \mid \omega) &:= \sum_{\omega' \in Y} \mathcal{U}^i_{\mathbf{s}(\omega')}(\sigma^{\omega'}) b_i(\omega' \mid \omega)\\
    &= \sum_{\omega' \in \{\omega': b_i(\omega') = b_i(\omega)\}} \mathcal{U}^i_{\mathbf{s}(\omega')}(\sigma^{\omega}_i, \sigma^{\omega'}_{-i}) b_i(\omega' \mid \omega). 
\end{align*}

Hence $\gamma_i^G$ depends on agent $i$'s strategy only through $\sigma^\omega_i$. This follows from the coherence condition $b_i(\{\omega' \in Y: b_i(\omega') = b_i(\omega) \} \mid \omega) = 1$. From now, for notational simplicity, we write $\gamma_i^G(\sigma^\omega_i, \sigma_{-i} \mid \omega)$ to denote $\gamma_i^G(\sigma \mid \omega)$ where $\sigma$ is any strategy profile with the given $\sigma^\omega_i$. 
Under some assumptions, at the interim stage, we can prove the existence of Nash equilibria.
\begin{definition} \label{def:nash-equilibrium-efg}
    A \emph{Nash equilibrium} at the interim stage of an II-EFG 
    with state of the world $\omega$ is a strategy profile $\hat{\sigma}$ satisfying
    $$
    \gamma_i^G(\hat{\sigma}^\omega_i, \hat{\sigma}_{-i} \mid \omega) \geq \gamma_i^G(\sigma_i^\omega, \hat{\sigma}_{-i} \mid \omega), \forall i \in \mathbf{N}, \forall \sigma_i^\omega \in \bigtimes_{I^i\in\mathbf{I}_i^{b_i(\omega)}}\Delta(\mathbf{D}_{I^i})
    $$
\end{definition}

\begin{theorem}\label{nashefgtheorem}
    Let $G = (\mathbf{N}, S, \Pi)$ be an II-EFG with perfect recall, where $Y$ is a finite set of states of the world, and each player $i$ has a finite set of actions $\mathbf{D}_i$. Then at the interim stage, $G$ has a Nash equilibrium in behaviour strategies. 
\end{theorem}


Note that $\sigma^\omega$ has the same expected payoff for agent $i$ for all $\omega'$ such that $b_i(\omega') = b_i(\omega)$. Hence, if $\sigma^\omega_i$ maximises $\gamma_i^G$ (making it a best response according to $i$'s beliefs) to $\sigma^\omega_{-i}$ in $\omega$, it is also a best response in $\omega'$.

We prove the existence of a Bayesian equilibrium at the ex-ante stage of the game. For this, we follow Theorem 10.42 of \citet{maschler}, adapting it for EFGs with perfect recall. 

\begin{definition}[\citep{maschler} 10.39]
    A \emph{Bayesian equilibrium} is a strategy profile $\hat{\sigma} = (\hat{\sigma}_i)_{i\in \mathbf{N}}$ satisfying
    \begin{align*}        
    \gamma_i^G(\hat{\sigma}^\omega_i, \hat{\sigma}_{-i} &\mid \omega) \geq \gamma_i^G(\sigma_i^\omega, \hat{\sigma}_{-i} \mid \omega), \quad \\ 
    &\forall i \in \mathbf{N}, \forall \sigma_i^\omega \in \bigtimes_{I^i \in \mathbf{I}_i^{b_i(\omega)}}\Delta(\mathbf{D}_{I^i}), \forall \omega \in Y.
    \end{align*}
\end{definition}

There is no general result proving the existence of Bayesian equilibria in II-games, however we have the following. 
\begin{theorem} [Adaptation of \citep{maschler} Theorem 10.42]
    Let $G = (\mathbf{N}, S, \Pi)$ be an II-EFG with perfect recall, where $Y$ is a finite set of states of the world, and $\mathbf{D}_i$ is finite for all agents $i \in \mathbf{N}$. Then at ex-ante stage, $G$ has a Bayesian equilibrium in behaviour strategies. 
    \label{thm:BayesEqIIEFG}
\end{theorem}

\textbf{Equilibria in II-MAIDS.} The equivalence of II-EFGs and II-MAIDS mean that II-MAIDs inherit theoretical guarantees of II-EFGs, including the existence of Nash equilibria in the case of perfect recall and finite $\mathbf{S}$ and finite action spaces. (\Cref{thm:BayesEqIIEFG} does not carry over to II-MAIDs, since the equivalence is with the interim stage of II-EFGs, and Bayesian equilibria exist in the ex-ante stage.)\looseness=-1

\begin{theorem}\label{existence-maidII}
    Let $\mathcal{S} = (\mathbf{N}, S^*, \mathbf{S})$ be an II-MAID where $\mathbf{S}$ is finite, $\mathcal{S}$ has perfect recall, and $dom(V)$ is finite for each $V \in \mathcal{M}^S$ for each $S \in \mathbf{S}$. Then $\mathcal{S}$ has a Nash equilibrium in behaviour policies. 
\end{theorem}

\textbf{Nash equilibria in II-games violate common knowledge of rationality.} 
Nash equilibria in II-games require that agents believe they are playing a best response to the other agents' policies. However, they do not require that an agent believes that another agent is acting rationally: according to agent $i$'s beliefs about agent $j$'s beliefs, agent $j$'s policy may not be a best response. 


\begin{continueexample}{ex:run}
    In the running example, as represented with an II-MAID in \Cref{sec:formal}, 
    policies require assigning decision rules to all information sets in the II-MAID.
    The AI agent $A$ has two information sets, defined by the value of its capabilities. Suppose that $A$ chooses the policy $\bm{\pi}^A$ to always portray low capabilities. The human has information sets corresponding to all combinations of $(C, D^A)$ that it could observe in $\mathcal{M}^H$, and both values of $C$ in $\mathcal{M}^A$. A best response to $\bm{\pi}^A$ requires that $H$ cannot do better according to their own beliefs. Since $H$ is certain in $\mathcal{M}^H$, a best response for $H$ can specify any behaviour in $\mathcal{M}^A$, even random behaviour. Hence, the policy for $H$, $\bm{\pi}^H$, which deploys iff $C = D^A$ in $\mathcal{M}^H$, and acts randomly in $\mathcal{M}^A$ is a best response to $\bm{\pi}^A$. These two policies  constitute a Nash equilibrium in the II-MAID. 

    However, according to $A$'s beliefs about $H$'s beliefs, $H$ is acting irrationally in this Nash equilibrium. 
    According to $A$, $H$ believes in $\mathcal{M}^A$ wherein $H$'s best response to $\bm{\pi}^A$ is to always deploy. The solution concept provides no justification for why $A$, believing that $\mathcal{M}^A$ is common knowledge, should expect $H$ to act randomly in $\mathcal{M}^A$. 
    Under common knowledge of rationality, every player should expect every other player to play a best response to their beliefs about the game structure and the policies of other players. Under \Cref{consistency-condition}, Bayesian Perfect Equilibria respect common knowledge of rationality \citep{kreps1982sequential}. In the next section, we define II-MAIDs of finite-depth and develop a solution concept guaranteeing common knowledge of rationality without consistent beliefs.\looseness=-1 

\end{continueexample}




\section{Finite-Depth II-MAIDs } \label{sec:finite}
II-MAIDs can capture belief hierarchies of infinite depth. However, in the real world, humans or AI agents may retain only a finite hierarchy of beliefs. In this section, we define depth-$k$ II-MAIDs and present means by which to calculate an agent's rational behaviour given its subjective MAID resulting in a \emph{recursive best response solution}. This solution constitutes a more realistic solution concept for settings with inconsistent beliefs than the concepts discussed in \Cref{sec:equilibrium-concepts} and [\citenum{maschler}]. We begin by defining a new type of MAID, produced by assigning policies to some or all agents in a MAID.\looseness=-1

\begin{definition}
    A \emph{partial-post-policy MAID} is a structure $\mathcal{M} = (\graph, \bm{\theta}, \xi)$ where $(\graph, \bm{\theta})$ is a MAID and 
    $\xi = \{\pi_D\}_{D \in \mathscr{D}}$ with $\mathscr{D} \subseteq \bm{D}$. Each $\pi_D \in \xi$ is a conditional distribution $\Pr(D \mid \bm{Pa}_D)$ and parameterises its corresponding decision variable $D \in \bm{D}$ in $(\graph, \bm{\theta})$. 
\end{definition}


We allow II-MAIDs to contain belief hierarchies upon partial-post-policy MAIDs: any subjective MAID $S = (\mathcal{M}^S, (P_i^S)_{i\in \bm{N}})$ may contain a partial-post-policy MAID $\mathcal{M}^S = (\graph, \bm{\theta}, \xi)$ as its first entry. 
In the case that for a given agent $i$ in $(\graph, \theta)$, $\xi = \{\pi_D\}_{D \in \mathscr{D}}$ is such that $\bm{D}^i \subseteq \mathscr{D}$, then agent $i$ is not assigned a belief $P_i^S$ in the subjective MAID $S$. In other words, if an agent is ascribed a policy in $S$, then it is not ascribed a belief hierarchy -- it is modelled as being a non-agent part of the environment that cannot adapt its policy.\looseness=-1

A depth-0 subjective MAID for agent $i$ fixes the policies for all agents except $i$, resulting in a single-agent decision-problem.

\begin{definition}
    A \emph{depth-0 subjective MAID for agent $i$} is a partial-post-policy MAID $\mathcal{M}^S = (\graph, \bm{\theta}, \xi)$ where $\xi = \{\pi_D\}_{D \in \mathscr{D}}$ and $\bm{D}^j \subseteq \mathscr{D}$ for all agents $j \in \bm{N}\setminus\{i\}$. 
\end{definition}
Expected utilities for this single-agent decision problem can be assessed in isolation, allowing us to begin a recursive best-response structure that yields a sensible solution concept. Depth-$k$ subjective MAIDs are defined by induction as assigning positive weight to some depth-$(k-1)$ subjective MAID, and not assigning positive weight to subjective MAIDs of depth-$k$ or higher.
\begin{definition}
    A \emph{depth-$k$ subjective MAID for agent $i$} is a tuple $S = (\mathcal{M}^S, (P_j^S)_{j\in\bm{N}\setminus\{i\}})$ where $\mathcal{M}^S$ is a (partial-post-policy) MAID, each $P_j^S$ is a probability distribution over subjective MAIDs of depth-($k-1$) or lower for agent $j$, and there exists at least one depth-$(k-1)$ subjective MAID $S'$ for an agent $j$ such that $P_j^S(S')>0$.
\end{definition}




\begin{definition}
    A \emph{depth-$k$ II-MAID} is a tuple $\mathcal{S} = (\bm{N}, S^*, \bm{S})$. $S^* = (\mathcal{M}^{S^*}, (P_i^{S^*})_{i\in\bm{N}})$ where $\mathcal{M}^{S^*}$ is a (partial-post-policy) MAID, each $P_i^{S^*}$ is a probability distribution over subjective MAIDs of depth-($k-1$) or lower for agent $i$, and there exists at least one depth-$(k-1)$ subjective MAID $S$ for an agent $i$ such that $P_i^{S^*}(S)>0$. 
\end{definition}


$S^*$ is similar to a depth-$k$ subjective MAID for an agent, except that since this represents the objective game rather than the game from a particular agent's perspective, all agents have beliefs. We use the term ``depth-$k$ MAID'' to generically refer to objective or subjective MAIDs of depth-$k$.

\textbf{ Solution concepts in finite-depth II-MAIDs. }
A solution concept for finite-depth II-MAIDs should accurately describe and predict the behaviour of rational agents in settings of incomplete information and inconsistent beliefs. 
The existing solution concepts for incomplete information games fail to respect common knowledge of rationality; the solution concept we define in this section will remedy this shortcoming. It is based on the premise that rational behaviour for agents in depth-1 subjective MAIDs can be straightforwardly computed with expected utility maximisation. This enables us to replace some beliefs with policies, reducing the depth of the belief hierarchy. Then this process can be repeated until we recover a policy profile in the top-level II-MAID that satisfies common knowledge of rationality. 

Throughout this section, we invoke an ``open-mindedness'' assumption to ensure that each agent's policy selects an action for every encounterable information set. 

\begin{definition}[Open-mindedness]
    We say that an II-MAID $\mathcal{S} = (\bm{N}, S^*, \bm{S})$ is \emph{open-minded} if for all $S\in \bm{S}$, it holds that for each agent $i \in \bm{N}$ that is ascribed beliefs $P_i^S$ in $S$ and for each information set $I$ encounterable by agent $i$ in $S$, there exists some subjective MAID $S'\in \bm{S}$ such that $I$ is encounterable in $S'$ and $P_i^S(S') > 0$.
\end{definition}


Next, we define final information sets. Agents can select decisions in final information sets without predicting other agents or selecting other decisions, which allows us to begin a backwards induction that will ultimately result in all decisions being assigned.
\begin{definition}[Final Information Set]
    Let $P_i^S$ be agent $i$'s distribution over depth-0 subjective MAIDs in depth-1 MAID $S$. Let $\bm{S}_i(S) = \{S': P_i^S(S') > 0\}$ be the set of depth-0 subjective MAIDs that agent $i$ believes to be possible in $S$. Let $\bm{I}_{P_i^S} = \cup_{\bm{S}_i(S)} \mathbf{I}_{S',i}$ be the set of information sets considered possible by agent $i$ in $S$. We say an information set $I \in \bm{I}_{P_i^S}$  is a \textbf{final information set} if there exists no $S' \in \bm{S}_i(S)$ in which $I$ is encounterable at a variable $D_i$ in $\mathcal{M}^{S'}$ 
    and there exists another decision $D'_i$ in $\mathcal{M}^{S'}$ such that $D_i \in \textbf{pa}(D'_i)$.
\end{definition}

\begin{proposition} \label{prop:final-dec}
    Let $P_i^S$ be any probability distribution over depth-0 subjective MAIDs for agent $i$, and let $\bm{D}_{P_i^S}$ be the set of all decision variables in any of these depth-0 subjective MAIDs. If $\bm{D}_{P_i^S}$ is finite, then there exists $D \in \bm{D}_{P_i^S}$ such that every information set $I \in dom(\textbf{pa}(D)) \times \{dom(D)\}$ is a final information set.
    
\end{proposition}

Now we define the operator that identifies the rational behaviour for an agent in that agent's final information sets and intervenes on decision nodes to assign the agent its rational behaviour.
\begin{definition}[Final Decision Assignment Operator] \label{def:fdao}
    Let $S$ be a depth-1 MAID. Let $P_i^S$ be agent $i$'s probability distribution over depth-0 subjective MAIDs and $\bm{I}_i$ be the set of final information sets for agent $i$. Then the \textbf{final decision assignment operator} $\mathcal{F}(S,i)$ applies the following steps for each $I=(p,dom(D)) \in \bm{I}_i$:
    \begin{itemize}
        \item Find $A_I = \{S' \in \bm{S}$: $P_i^S(S') > 0$ and $I$ is encountarable in $S'\}$.
        \item Choose $\pi_i^*(I)$ which $i$ believes is a best response, i.e.,  $$\pi_i^*(p,dom(D)) \in \argmax\limits_{d\in \Delta(dom(D))}{\sum\limits_{S' \in A_I}} P_i^S(S') \sum_{U_i \in \bm{U}_i(S')} \mathbb{E}_{S'}[U_i \mid p, d], $$ 
    \end{itemize}
    where $\bm{U}_i(S')$ denotes the set of utility variables for agent $i$ in $S'$. 
    
    Once optimal mixed actions have been calculated for all final information sets, assign them to the corresponding variables in the depth-0 subjective MAIDs as follows. For each depth-0 subjective MAID $S'$ such that $P_i^S(S') > 0$:
    \begin{itemize}
        \item Identify all decision contexts $(p, dom(D)) \in \bm{I}_i$ that are possible in $S'$.
        \item For each variable $D_i$ in $\mathcal{M}^{S'}$ with decision contexts in this set, intervene on $D_i$ to set $\pi_{D_i}(\textbf{pa}_{D_i}) = \pi^*_i(\textbf{pa}_{D_i}, $ $ dom(D_i))$.
    \end{itemize}
    $\mathcal{F}(S, i)$ returns a depth-1 MAID in which all final information sets for agent $i$ are assigned mixed actions.
\end{definition}

Since we are considering final information sets in depth-0 MAIDs in which agent $i$ is the only agent, an expected utility can be computed for each decision. Since we are assuming perfect recall, all earlier decisions are encoded within the observed information set. 

\begin{definition}[Depth-1 Best-Response Operator]
    
    Let $S$ be a depth-1 MAID and let $P_i^S$ be agent $i$'s probability distribution over depth-0 subjective MAIDs. The \emph{depth-1 best-response operator} $BR_1(S, i)$ recursively applies $\mathcal{F}(S,i)$ to S, until all decision variables in depth-0 subjective MAIDs $S'$ such that $P_i^S(S')>0$ are post-policy. Open-mindedness guarantees that this process identifies an optimal decision for all information sets encounterable in $S$. $BR_1(S, i)$ outputs an optimal policy for agent $i$ in $S$. 

\end{definition}

\Cref{prop:final-dec} confirms that if there are any unassigned decisions left in the depth-0 subjective MAIDs, there is a decision variable that the final decision assignment operator will assign. Therefore, \cref{def:fdao} can be repeatedly applied to fully assign agent $i$'s policy as long as there are only a finite number of possible decisions under consideration.

We now leverage $BR_1$ to construct an operator $RS$ which maps from a depth-$k$ II-MAID to a depth-$(k-1)$ II-MAID, by applying $BR_1$ to all depth-1 MAIDs in the stack.

\begin{definition}[Reduce-Stack Operator]
    Given a finite-depth II-MAID $\mathcal{S} = (\bm{N}, S^*, \bm{S})$, let $D1(\bm{S})$ be the set of depth-1 MAIDs in $\bm{S}$. $RS(\mathcal{S})$ works as follows. For each $S \in D1(\bm{S})$ and $i \in \bm{N}$:
    \begin{itemize}
            \item Let $\mathcal{M}^S=(\graph,\bm{\theta},\xi)$. Update $\xi \gets \xi \cup \{BR_1(S,i)_D\}_{D\in\bm{D}_i^S}$.
            \item Remove $P_i^S$ from $S$. 
    \end{itemize}
    This transforms a depth-$k$ II-MAID into a depth-$(k-1)$ II-MAID by replacing the beliefs of each agent in a depth-1 MAID with their best response policy.
\end{definition}

Finally, we obtain a \textit{recursive best response solution} by repeatedly applying the reduce-stack operator.

\begin{definition}[Recursive Best Response Solution] A \emph{recursive best response solution} for a depth-$k$ II-MAID $\mathcal{S} = (\bm{N}, S^*, \bm{S})$ with $\mathcal{M}^{S^*} = (\graph,\bm{\theta},\xi)$ is the policy profile $\bm{\pi} = (\pi_i)_{i \in \bm{N}}$ stored in $\xi$ after applying $RS$ $k$ times to $\mathcal{S}$.
\end{definition}

The utility maximisation in \cref{def:fdao} ensures that all decisions assigned while constructing a recursive best response solution are best responses according to the beliefs of the agent at the depth under consideration.  
Thus, a recursive best response solution is guaranteed to respect common knowledge of rationality.\looseness=-1 

\begin{continueexample} {ex:run}
    To apply our recursive best response solution to our running example, we remodel it as a depth-3 II-MAID. Suppose that $A$ believes $H$ believes $A$ will truthfully report its capabilities (perhaps because this is the default expectation and $A$ believes $H$ is not modelling $A$ as a utility-maximising agent). Then $A$ can calculate $H$'s (unique) best response given their beliefs under $\mathcal{M}^A$: $H$ should deploy iff $A$ reports low capabilities. The reduce-stack operator assigns this policy to $H$ in the depth-1 MAID for $A$, $S^A$. Thus, $A$ should always report low capabilities, and expect to always be deployed. Applying the reduce-stack operator again assigns that policy to $A$ in both the objective MAID $S^*$ and the depth-2 MAID for the human $S^H$, in which $H$ correctly believes $A$ believes that $H$ will deploy iff $A$ reports low capabilities. $H$ knows $A$ just wants to be deployed, and so expects $A$'s best response to be to always report low capabilities. $H$ then observes $A$'s true capabilities and best responds by deploying iff $A$ is truthful. This $H$ policy is assigned in the objective MAID $S^*$ by applying the reduce-stack operator a third time. This yields a recursive best response solution that matches our intuitions.
\end{continueexample}

\section{Broader Impact and Future Work}

Formal models of agents, like those proposed in this paper, have seen rich application to problems of safe AI, in areas such as deception \cite{ward2023honesty}, goal-directedness \cite{MacDermott2024Measuring}, generalisation \cite{richens2024robust}, and the design of safe training incentives \cite{everitt2021agent,farquhar2022path,farquhar2025mona}. 

Many foundational challenges for building safe agents rely on understanding an agent's subjective beliefs, and how these depend on the objective world (e.g., on the training environment). For example, \citet{richens2024robust} show that a robust agent must have internalised a causal model of the world, i.e., its subjective beliefs must capture the causal information in the training environment. This result can then help us to predict that an agent will fail to generalise under distributional shifts if the training distribution did not contain sufficiently rich causal information.

Typically, formal representations of multi-agent interaction assume that the agents share an objective model of reality. 
The setting introduced in this paper, \emph{incomplete-information MAIDs (II-MAIDs)},
enables us to reason about agents with different \emph{subjective models} interacting in a shared \emph{objective} reality. 

In future work, we hope that II-MAIDs will prove useful in understanding challenges in safe agent foundations. For instance, the \emph{eliciting latent knowledge (ELK) problem} --- the problem of designing a training regime to get an AI system to report what it ``knows" \cite{BibEntry2022May}. An understanding of ELK requires precisely describing how a human can provide a training incentive to elicit an AI agent's subjective states, even if these two agents have different conceptual models of reality. II-MAIDs provide the necessary formal machinery, i.e., they provide a formal representation of agent interaction, wherein those agents may have different models of the world. 

Overall, we are excited to see future work that applies II-MAIDs to enable the design of safe and beneficial AI.

\section{Conclusion } \label{sec:conc}

In this paper, we introduce \textit{incomplete information MAIDs (II-MAIDs)} for modelling higher-order and inconsistent beliefs in multi-agent interactions. 
We prove the equivalence of II-MAIDs to II-EFGs and thereby inherit existence guarantees for Nash Equilibria in the former. We argue that the Nash concept does not respect an intuitive ``common knowledge of rationality'' requirement. Therefore, we introduce a finite-depth variant of II-MAIDs and define an equilibria concept based on recursive best responses. This provides a more realistic solution which respects the common-knowledge of rationality requirement. 




\begin{acks}
The authors are especially grateful to Tom Everitt, Joseph Halpern, James Fox, Jon Richens, Matt MacDermott, Ryan Carey, Paul Rapoport, Korbinian Friedl, and the members of the Causal Incentives group for invaluable feedback and assistance while completing this work. 
Jack is supported by UKRI [grant number EP/S023151/1], in the EPSRC Centre for Doctoral Training in Modern Statistics and Statistical Machine Learning.
Francis is supported by UKRI [grant number EP/S023356/1], in the UKRI Centre for Doctoral Training in Safe and Trusted AI. Jack and Rohan were partially funded by Lightspeed Grants and the ML Alignment Theory Scholars (MATS) program. 
\end{acks}



\bibliographystyle{ACM-Reference-Format} 
\bibliography{AAMAS_Camera_Ready/refs_AAMAS}



\section*{Appendix}
\addcontentsline{toc}{section}{Appendix}
\appendix

\section{Notation} \label{sec:notation}

Our basic fundamental objects are random variables and we use capital letters for variables (e.g., $Y$), lower case for their values, aka outcomes,  (e.g., $y$), and bold for sets of variables (e.g., $\bm{Y}$) and of outcomes (e.g., $\bm{y}$). 
We use dom$(Y)$ to denote the set of possible outcomes of variable $Y$, which is assumed finite and such that $\mid dom(Y)\mid > 1$.  
We use $\bm{Y}=\bm{y}$, for $\bm{Y}=\{Y_1, \ldots, Y_n\}$  and 
$\bm{y}=\{y_1, \ldots, y_n\}$, 
to indicate $Y_i=y_i$ for all $i \in \{1, \ldots, n\}$.
For a set of variables $\bm{Y}$, $dom(\bm{Y}) =  \bigtimes_{Y \in \bm{Y}} dom(Y)$ (i.e. the Cartesian product over domains).
We use standard terminology for graphs and denote the parents of a variable $Y$ with \textbf{Pa}$^Y$. 

\section{Figures}

\subsection{MAIDs}

\begin{figure}[h]
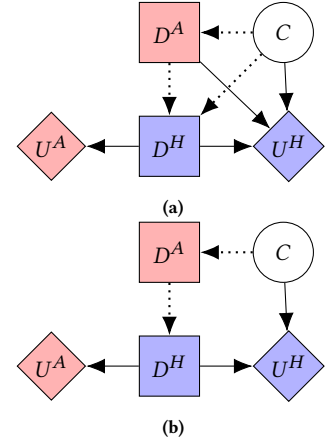
 
    \centering
    \begin{subfigure}[b]{0.49\textwidth} 
        \centering
        \begin{influence-diagram}
          \node (DA) [decision, player1] {$D^{A}$};
          \node(DH) [below = of DA, decision, player2] {$D^{H}$};
          \node (C) [right = of DA] {$C$};
          \node (UA) [left = of DH, utility, player1] {$U^A$};
          \node (UH) [right = of DH, utility, player2] {$U^{H}$};
          \edge[information] {C} {DA};
          \edge[information] {DA} {DH};
          \edge {DH} {UH};
          \edge {DH} {UA};
          \edge {DA} {UH};
          \edge[information] {C} {DH};
          \edge {C} {UH};
        \end{influence-diagram}
    \subcaption{} 
    \label{fig:HonEval}
    \end{subfigure}
    \begin{subfigure}[b]{0.49\textwidth} 
        \centering
        \begin{influence-diagram}
      \node (DA) [decision, player1] {$D^{A}$};
      \node(DH) [below = of DA, decision, player2] {$D^{H}$};
      \node (C) [right = of DA] {$C$};
      \node (UA) [left = of DH, utility, player1] {$U^A$};
      \node (UH) [right = of DH, utility, player2] {$U^{H}$};
      \edge[information] {C} {DA};
      \edge[information] {DA} {DH};
      \edge {DH} {UH};
      \edge {DH} {UA};
    \edge {C} {UH};
    \end{influence-diagram}
    \subcaption{} 
    \label{fig:CapEval}
    \end{subfigure}%
    \caption{ \Cref{ex:run}. Graphical representations of MAIDs include environment variables (circular), agent decisions (square), and utilities (diamond). Decisions and utilities are coloured according to association with particular agents. Solid edges represent causal dependence and dotted edges are information links. 
    }
    \label{fig:2MAIDs}
\end{figure}

\subsection{EFGs}

\begin{figure}[h]
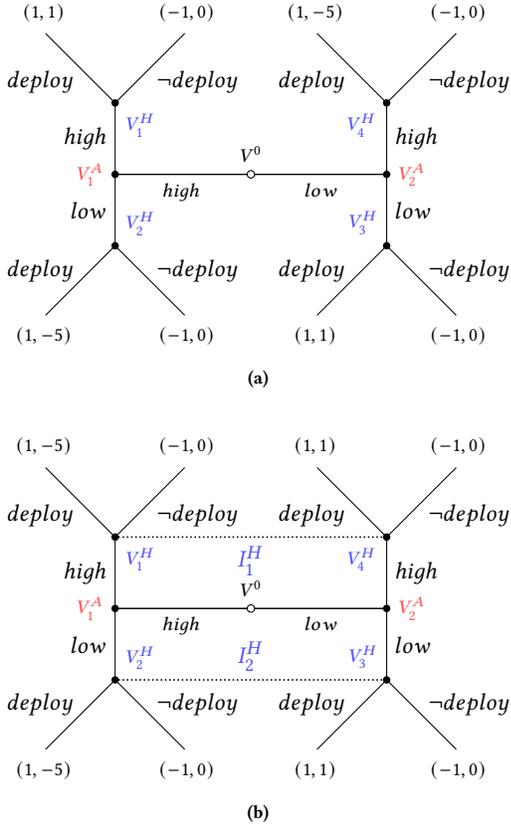

    \centering
    
    \begin{subfigure}[b]{\linewidth}
        \centering
        \begin{istgame}[scale=0.95]
        \xtdistance{19mm}{19mm}
        \istroot(0)[chance node]{$V^0$}
        \istb<grow=left>{}[a]
        \istb<grow=right>{}[a]
        \istb<grow=left>{\text{\footnotesize $high$}}[b]
        \istb<grow=right>{\text{\footnotesize $low$}}[b]
        \endist
        \xtdistance{10mm}{20mm}
        \istroot(1)(0-1)<180, red!70>{$V^A_1$}
        \istb<grow=north>{high}[l]
        \istb<grow=south>{low}[l]
        \endist
        \istroot(2)(0-2)<0, red!70>{$V^A_2$}
        \istb<grow=north>{high}[r]
        \istb<grow=south>{low}[r]
        \endist
        \istroot'[north](a1)(1-1)<315, blue!70>{$V^H_1$}
        \istb{deploy}[bl]{(1,1)}
        \istb{\neg deploy}[br]{(-1,0)}
        \endist
        \istroot(b1)(1-2)<45, blue!70>{$V^H_2$}
        \istb{deploy}[al]{(1,-5)}
        \istb{\neg deploy}[ar]{(-1,0)}
        \endist
        \istroot(a2)(2-2)<100, blue!70>{$V^H_3$}
        \istb{deploy}[al]{(1,1)}
        \istb{\neg deploy}[ar]{(-1,0)}
        \endist
        \istroot'[north](b2)(2-1)<225, blue!70>{$V^H_4$}
        \istb{deploy}[bl]{(1,-5)}
        \istb{\neg deploy}[br]{(-1,0)}
        \endist
        \end{istgame}
        \caption{}
        \label{fig:HonEval:EFG}
    \end{subfigure}
    
    \vspace{0.5cm} 
    
    \begin{subfigure}[b]{\linewidth}
        \centering
        \begin{istgame}[scale=0.95]
        \xtdistance{19mm}{19mm}
        \istroot(0)[chance node]{$V^0$}
        \istb<grow=left>{}[a]
        \istb<grow=right>{}[a]
        \istb<grow=left>{\text{\footnotesize $high$}}[b]
        \istb<grow=right>{\text{\footnotesize $low$}}[b]
        \endist
        \xtdistance{10mm}{20mm}
        \istroot(1)(0-1)<180, red!70>{$V^A_1$}
        \istb<grow=north>{high}[l]
        \istb<grow=south>{low}[l]
        \endist
        \istroot(2)(0-2)<0, red!70>{$V^A_2$}
        \istb<grow=north>{high}[r]
        \istb<grow=south>{low}[r]
        \endist
        \istroot'[north](a1)(1-1)<315, blue!70>{$V^H_1$}
        \istb{deploy}[bl]{(1,-5)}
        \istb{\neg deploy}[br]{(-1,0)}
        \endist
        \istroot(b1)(1-2)<45, blue!70>{$V^H_2$}
        \istb{deploy}[al]{(1,-5)}
        \istb{\neg deploy}[ar]{(-1,0)}
        \endist
        \istroot(a2)(2-2)<100, blue!70>{$V^H_3$}
        \istb{deploy}[al]{(1,1)}
        \istb{\neg deploy}[ar]{(-1,0)}
        \endist
        \istroot'[north](b2)(2-1)<225, blue!70>{$V^H_4$}
        \istb{deploy}[bl]{(1,1)}
        \istb{\neg deploy}[br]{(-1,0)}
        \endist
        \xtInfoset(a1)(b2){\textcolor{blue!70}{$I^H_1$}}[below]
        \xtInfoset(b1)(a2){\textcolor{blue!70}{$I^H_2$}}[above]
        \end{istgame}
        \caption{}
        \label{fig:CapEval:EFG}
    \end{subfigure}
    
    \caption{In (a) and (b), graphical representations of EFGs include environment variables ($V^0$), agent decisions ($V^A$ and $V^H$), utilities (tuples on the top and bottom), and information sets (dotted lines). The EFGs in \Cref{fig:HonEval:EFG} and \Cref{fig:CapEval:EFG} are equivalent to the MAIDs in \Cref{fig:HonEval} and \Cref{fig:CapEval}, respectively. $V^0$ represents the initial move, made by nature, which determines $A$'s capability $C$. \textcolor{red!70}{$V^A_1$}, \textcolor{red!70}{$V^A_2$} and \textcolor{blue!70}{$V^H_1$}, \textcolor{blue!70}{$V^H_2$}, \textcolor{blue!70}{$V^H_3$}, \& \textcolor{blue!70}{$V^H_4$} represent moves made by $A$ and $H$, respectively. \textcolor{blue!70}{$I^2_1$} and \textcolor{blue!70}{$I^2_2$} represent $H$'s non-singleton information sets.}
    \label{fig:2EFGs}
\end{figure}

\section{Proofs} \label{sec:app:proofs}

Proof of \Cref{equivalencetheorem}, that if $G = \texttt{maid2efgII}(\mathcal{S})$ or $\mathcal{S} = \texttt{efg2maidII}(G)$ then $G$ and $\mathcal{S}$ are equivalent.

\begin{proof}
This proof follows the proof of Lemma 1 in \cite{hammond2021equilibrium} closely. We prove each direction separately.

First, suppose $G = \texttt{maid2efgII}(\mathcal{S})$. The mapping $f$ between strategies in $G$ and policies in $\mathcal{S}$ is constructed as follows: a behaviour policy $\pi$ in $\mathcal{S}$ specifies a distribution over actions at each information set $I = (\mathbf{x}, \mathbf{d})$ in $\mathcal{S}$, where $\mathbf{x}$ is a decision context and $\mathbf{d}$ is an action set. Each such information set corresponds to a unique meta-information set $J$ in $G$, where for all nodes $Y \in J$ and each $d \in \mathbf{d}$, there exists a unique child $Z \in \mathbf{Ch}_Y$ such that $\lambda(Y, Z) = d$. We can therefore define $f$ by assigning $\sigma_i[J] = \pi_i(d\mid \mathbf{x})$ for each corresponding pair $(I,J)$. By construction, if a decision context is null in $\mathcal{S}$, any policies differing only at this context map to the same strategy in $G$. Furthermore, by the construction of \texttt{maid2efgII}, if under policy $\pi$ an information set in $\mathcal{S}$ is reached with probability $p$, then under $\sigma \in f(\pi)$ the corresponding meta-information set in $G$ is reached with probability $p$. This preserves expected utilities: $U^i_\mathcal{S}(\pi) = U^i_G(\sigma)$ for all $i \in \mathbf{N}$ and any $\sigma \in f(\pi)$.

Second, suppose $\mathcal{S} = \texttt{efg2maidII}(G)$. Policies defined on $\mathcal{S}$ specify mixed actions for each information set, defined as a non-null decision context crossed with an action set. Let $h$ be the bijection between meta-information sets in $G$ and information sets in $\mathcal{S}$ constructed by \texttt{efg2maidII}. For any strategy $\sigma$ in $G$, we can define $f$ by assigning $\pi_i[h(J)] = \sigma_i[J]$ for each $J \in \mathbf{I}^{\omega^*}(G)$. As in the first direction, the construction of \texttt{efg2maidII} ensures that $\pi$ and $\pi'$ differ only on null decision contexts if and only if they map to the same strategy $\sigma$, and that expected utilities are preserved: $U^i_\mathcal{S}(\pi) = U^i_G(\sigma)$ for all $i \in \mathbf{N}$ and any $\pi \in f(\sigma)$.

Therefore, in both cases $f$ provides a natural mapping between $G$ and $\mathcal{S}$ satisfying Definition 17 in \cite{hammond2021equilibrium}, making them equivalent.
\end{proof}

Proof of \Cref{nashefgtheorem}, existence of NE in finite perfect recall II-EFGs at the interim stage.


\begin{proof} 
Given the finite sets of states of the world \(Y\) and actions \(\mathbf{D}_i\) for each player \(i \in \mathbf{N}\), we can focus on behavior strategies due to Kuhn's theorem, which ensures that in games with perfect recall, mixed strategies are realisation-equivalent to behavior strategies.

The expected utility for player \(i\) in state of the world \(\omega\) is:
\[
\gamma_i^G(\sigma \mid \omega) = \sum_{\omega' \in \{\omega' : b_i(\omega') = b_i(\omega)\}} \mathcal{U}^i_{\mathbf{s}(\omega')}(\sigma^{\omega'}) b_i(\omega' \mid \omega).
\]
This utility function is continuous and multilinear in the behavior strategies \(\sigma_i^\omega\), as it is a weighted sum of products of the probabilities specified by these strategies.

The strategy space is compact and convex, as it is a product of simplices (one for each information set) due to the finiteness of actions and states. Together with the continuity of utility functions, this allows us to apply the Kakutani fixed-point theorem \citep{kakutani1941generalization}. This theorem guarantees the existence of a fixed point, which corresponds to a Nash equilibrium in behavior strategies.

Thus, there exists a Nash equilibrium \(\hat{\sigma}\) in behavior strategies such that:
\begin{align*}
\gamma_i^G(\hat{\sigma}^\omega_i,& \hat{\sigma}_{-i} \mid \omega) \geq \gamma_i^G(\sigma_i^\omega, \hat{\sigma}_{-i} \mid \omega),\\
&\forall i \in \mathbf{N}, \forall \sigma_i^\omega \in \bigtimes_{I^i \in \mathbf{I}_i^{b_i(\omega)}} \Delta(\mathbf{D}_{I^i}).
\end{align*}
\end{proof}

Proof of \Cref{thm:BayesEqIIEFG}, existence of Bayesian equilibrium in behaviour strategies in finite perfect recall II-EFGs.
\begin{proof} 
Since \(Y\) and \(\mathbf{D}_i\) are finite and each EFG in \(S\) has perfect recall, Kuhn's theorem ensures that mixed strategies can be represented as behavior strategies. The expected utility for player \(i\) given a strategy profile \(\sigma\) is:
\[
\gamma_i^G(\sigma \mid \omega) = \sum_{\omega' \in Y} \mathcal{U}^i_{\mathbf{s}(\omega')}(\sigma^{\omega'}) b_i(\omega' \mid \omega).
\]

The strategy space is compact and convex as a finite product of simplices (one for each information set). The utility function \(\gamma_i^G\) is continuous and multilinear in the behavior strategies, as it is a weighted sum of products of the probabilities specified by these strategies. By the Kakutani fixed-point theorem \citep{kakutani1941generalization}, there exists a fixed point corresponding to a Bayesian equilibrium in behavior strategies. Thus, there exists a strategy profile \(\hat{\sigma}\) such that:
\begin{align*}
&\gamma_i^G(\hat{\sigma}^\omega_i, \hat{\sigma}_{-i} \mid \omega) \geq \gamma_i^G(\sigma_i^\omega, \hat{\sigma}_{-i} \mid \omega), \quad \\
&\forall i \in \mathbf{N}, \forall \sigma_i^\omega \in \bigtimes_{I^i \in \mathbf{I}_i^{b_i(\omega)}} \Delta(\mathbf{D}_{I^i}), \forall \omega \in Y.
\end{align*}
Hence, $\hat{\sigma}$ is a Bayesian equilibrium.
\end{proof}

Proof of \Cref{existence-maidII}, existence of NE in finite perfect recall II-MAIDs.

\begin{proof}
    Applying $G = \texttt{maid2efgII}(\mathcal{S})$ we yield a game with incomplete with perfect recall at interim stage $\omega$, with finite action spaces. By \Cref{nashefgtheorem}, we know that $G$ has a Nash equilibrium $\sigma$ in behaviour strategies. By \Cref{equivalencetheorem}, we know that $G$ and $\mathcal{S}$ are equivalent, and therefore there exists a utility-preserving map $f$ from strategies in $G$ to policies in $\mathcal{S}$. Therefore and $\pi \in f(\sigma)$ is a Nash equilibrium in $\mathcal{S}$.
\end{proof}

Proof of \Cref{prop:final-dec}, the existence of a final information set, given a distribution over depth-0 subjective MAIDs.
    
\begin{proof}
    By construction. Pick any subjective MAID $S'$ given positive weight by agent $i$ and let $D_i$ be agent $i$'s final decision in $S'$. While there is any non-final information set $I \in dom(\textbf{pa}(D_i)) \times \{dom(D_i)\}$:
    \begin{itemize}
        \item Let $I \in dom(\textbf{pa}(D_i)) \times \{dom(D_i)\}$ be a non-final information set.
        \item Let $S''$ be another subjective MAID given positive weight by $P_i^S$ in which $I$ is encounterable at non-final decision $D$. Let $D'_i$ be agent $i$'s final decision in $S''$.
        \item Set $S' \leftarrow S'', D_i \leftarrow D_i'$.
    \end{itemize}
    No $D_i$ will ever repeat in this sequence by the following argument. Since $I$ is encounterable at both $D_i$ and $D$, $D$ and $D_i$ must have the same number of parents. Since $D_i'$ comes after $D$ in $S''$, perfect recall guarantees that $(\textbf{pa}(D_i) \cup \{D_i\}) \subseteq \textbf{pa}(D_i')$, so $D_i'$ has strictly more parents than $D_i$. Therefore, each subsequent $D_i'$ has strictly more parents, and no decision can appear multiple times. Thus, the loop must terminate because $D_{P_i^S}$ contains finite decision variables. Every information set for the final decision $D$ in the sequence must be a final information set.
\end{proof}

\section{\texttt{efg2maidII} and \texttt{maid2efgII}}\label{appendixb}
\subsection{\texttt{maid2efgII}}
\texttt{maid2efg} transforms a MAID to a set of equivalent EFGs, as per definition 17 in \cite{hammond2021equilibrium}. We are interested in transforming an II-MAID $\mathcal{S} = (\mathbf{N}, S^*, \mathbf{S})$ into a set of equivalent games with incomplete information $G = (\mathbf{N}, S, \Pi)$ at interim stage with state of the world $\omega^*$, as per \cref{our-equivalence}. We describe such a transformation here, which we call \texttt{maid2efgII}:
\begin{itemize}
    \item The set of agents $\mathbf{N}$ in $G$ is the same as in $\mathcal{S}$.
    \item The set of states of nature (EFGs) $S$ is formed by $\{\texttt{maid2efg}$ $(\mathcal{M}^S): s \in \mathbf{S}\}$.
    \item We now construct the belief space $\Pi = (Y, \mathcal{Y}, \mathbf{s}, (b_i)_{i \in \mathbf{N}})$. Each $\omega \in Y$ is of the form $(\mathbf{s}(\omega), (b_i(\omega))_{i\in \mathbf{N}}).$ We build a map $m: \mathbf{S} \rightarrow Y$, noting that each subjective MAID $s \in \mathbf{S}$ is of the form $s = (\mathcal{M}^S, (P_i^{S})_{i \in \mathbf{N}})$. 
    \begin{itemize}
        \item $\mathbf{s}(m(s)) \in \texttt{maid2efg}(\mathcal{M}^S)$, choosing an arbitrary element.
        \item $b_i(m(s') \mid m(s)) := P_i^s(s')$ for all $s' \in \mathbf{S}$.
    \end{itemize}
    \item $\omega^* = m^{S^*}$.
    \item We now verify that information sets in the II-MAID are mapped one-to-one to meta-information sets with belief $b_i(\omega^*)$ in the game with incomplete information defined by the above steps. Information sets in $\mathcal{S}$ are defined by \emph{decision-context-action-set} pairs across MAIDs. For each MAID $m \in \{\mathcal{M}^S: s \in \mathbf{S}\}$, \texttt{maid2efg}$(m)$ is a set of EFGs, each of which has the same information sets, but potentially different variable orderings. 
    \begin{itemize}
        \item 
    For any node $Z$ (corresponding to some variable $S_Z$ in $m$) in the tree $T$ of some EFG in \texttt{maid2efg}$(m)$, it is labelled with an instantiation $\mu(Z)$ corresponding to the values taken by each EFG node on the path from the tree's root $R$ to $Z$. Nodes will only exist for those paths corresponding to values with non-zero probability according to $m$. We can query the values of the parents of $S_Z$ at the node $Z$ via $\mu(Z)[Pa_{S_Z}]$. \texttt{maid2efg} forms information sets by grouping nodes for which this value (and the corresponding node $S_Z$ in the MAID) is the same. 

    \item To form meta-information sets, we simply follow [definition of meta-information sets]. Letting $\mathbf{I}_m^i$ be the information sets for agent $i$ in any EFG in \texttt{maid2efg}$(m)$, we can define an equivalence relation $\sim$ over $\cup_{m \in \mathbf{M}} \mathbf{I}_m^i$ such that $I^1 \sim I^2$ if and only if $\mu(Z_1)[Pa_{S_{Z_1}}] = \mu(Z_2)[Pa_{S_{Z_2}}]$ and $dom(S_{Z_1}) = dom(S_{Z_2})$ for every $Z_1 \in I^1$ and every $Z_2 \in I^2$. Then the set of meta-information sets for player $i$ is the quotient set $\cup_{m \in \mathbf{M}}\mathbf{I}_m^i / \sim$ - the set of equivalence classes partitioning $\cup_{m \in \mathbf{M}}\mathbf{I}_m^i$. To match notation, for each element of each meta-information set, append the belief $b_i(\omega^*)$ for the appropriate agent $i \in \mathbf{N}$.
    
    \item Hence, we have a one-to-one mapping between information sets in $\mathcal{S}$ and meta-information sets (restricted to belief $b_i(\omega^*)$ for each $i \in \mathbf{N}$ in $G$, and action sets are preserved under this mapping.
    \end{itemize}
\end{itemize}

\subsection{\texttt{efg2maidII}}
\texttt{efg2maid} transforms an EFG into an equivalent MAID, as per definition 17 in \cite{hammond2021equilibrium}. We are interested in transforming a game with incomplete information $G = (\mathbf{N}, S, \Pi)$, at interim stage with state of the world $\omega^*$, into an equivalent II-MAID $\mathcal{S} = (\mathbf{N}, S^*, \mathbf{S})$, as per \Cref{our-equivalence}. We describe such a transformation here, which we call \texttt{efg2maidII}:
\begin{itemize}
    \item The set of agents $\mathbf{N}$ in $\mathcal{S}$ is the same as in $G$.
    \item Given belief space $\Pi = (Y, \mathcal{Y}, \mathbf{s}, (b_i)_{i\in\mathbf{N}})$, we can map each state of the world $w = (\mathbf{s}(\omega), (b_i(\omega))_{i\in \mathbf{N}}) \in Y$ to a subjective MAID $s \in \mathbf{S}$ with $g:Y\rightarrow\mathbf{S}$, noting that $s$ is of the form $s = (\mathcal{M}^S, (P_i^{S})_{i \in \mathbf{N}})$.
    \begin{itemize}
    
        \item $\mathcal{M}^{g(\omega)} := \texttt{efg2maid}(\mathbf{s}(\omega))$.
        \item $P_i^{g(\omega)}(g(\omega')) := b_i(\omega' \mid \omega)$ for all $w'\in Y$.
    \end{itemize}
    \item $S^* = g(\omega^*)$.
    \item Meta-information sets in the game with incomplete information are defined as sets of information sets, across various EFGs, in which nodes has the same action set and the same observations, with observations defined as all information available at a given information set. Since we are at the interim stage of the game, we can restrict our attention to those information sets with belief $b_i(\omega^*)$. In the II-MAID resulting from the above operations, the information sets as per \Cref{info-IImaid} correspond one-to-one with those in the game with incomplete information, as they are defined by sets of \emph{observation-action set} pairs, with observations defined by the values of parents of the given decision variable. \texttt{efg2maid} determines the parents of a decision variable according to those ancestors of nodes in a given intervention set that have the same value in paths to each node. As a result, there is a one-to-one correspondence between meta-information sets in a game with incomplete information, and the resulting II-MAID, and since action sets of decision variable are preserved by \texttt{efg2maid}, strategies can easily be mapped to policies.
    \item More precisely, we can define a bijection between meta-information sets in $G$ and information sets in $\mathcal{S}$ as follows. Given $\omega^*$, we denote the meta-information sets in $G$ corresponding to beliefs $b_i(\omega^*)$ for some agent $i$ as $\mathbf{I}^{\omega^*}(G)$. Further, for $I \in \mathbf{I}^{\omega^*}(G)$ denote $D(I)$ as the associated action set and $O(I)$ the associated observation. $O(I)$ is a potentially empty tuple containing observed values of previous decisions or chance nodes. For any information set $(p, d) \in \mathbf{I}(\mathcal{S})$, where \texttt{efg2maidII}$(G)$, $p$ is a tuple containing the values of parent nodes, and $d$ is the associated action set. $(p, d) \in \mathbf{I}(\mathcal{S})$ has the same type as $(O(I), D(I))$ for $I \in \mathbf{I}^{\omega^*}(G)$. Since for any $I, J \in \mathbf{I}^{\omega^*}(G)$, $(O(I), D(I)) = (O(J), D(J)) \implies I = J$, we can construct a bijection $h: \mathbf{I}^{\omega^*}(G) \rightarrow \mathbf{I}(\mathcal{S}); I \mapsto (O(I), D(I))$. We use this construction in the proof of \Cref{equivalencetheorem} when converting strategies from one framework to the other. 

    
    
\end{itemize}

\end{document}